\documentclass[conference]{IEEEtran}
\IEEEoverridecommandlockouts
\usepackage{packages}
\DeclareRobustCommand*{\IEEEauthorrefmark}[1]{%
	\raisebox{0pt}[0pt][0pt]{\textsuperscript{\footnotesize #1}}%
}
\begin{document}
\graphicspath{{./images/}{images/datFiles/}{images/HRF/}{images/real-2d/}{images/real-3d/}{images/retina_2D/}{images/syn-2d/}{images/vascular-3d/}}	
\title{Curvilinear Structure Enhancement by Multiscale Top-Hat Tensor in 2D/3D Images}
\author{
	\IEEEauthorblockN{Shuaa S. Alharbi\IEEEauthorrefmark{1,}\IEEEauthorrefmark{2}, \c{C}i\u{g}dem~Sazak\IEEEauthorrefmark{1}, Carl~J.~Nelson\IEEEauthorrefmark{3}, Boguslaw Obara\IEEEauthorrefmark{1,}\IEEEauthorrefmark{*}}
	\IEEEauthorblockA{\IEEEauthorrefmark{1} Department of Computer Science, Durham University, Durham, UK
	\\\{shuaa.s.alharbi,cigdem.sazak,boguslaw.obara\}@durham.ac.uk}
	 \IEEEauthorblockA{\IEEEauthorrefmark{2} Computer College, Qassim University, Qassim, KSA
	 	\\shuaa.s.alharbi@qu.edu.sa}
	\IEEEauthorblockA{\IEEEauthorrefmark{3}School of Physics and Astronomy, University of Glasgow, Glasgow, UK
		\\chas.nelson@glasgow.ac.uk}
	\thanks{$^{*}$ Corresponding author: boguslaw.obara@durham.ac.uk
	}
}
	\maketitle
\begin{abstract}
A wide range of biomedical applications require enhancement, detection, quantification and modelling of curvilinear structures in 2D and 3D images. Curvilinear structure enhancement is a crucial step for further analysis, but many of the enhancement approaches still suffer from contrast variations and noise. This can be addressed using a multiscale approach that produces a better quality enhancement for low contrast and noisy images compared with a single-scale approach in a wide range of biomedical images. Here, we propose the Multiscale Top-Hat Tensor (MTHT) approach, which combines multiscale morphological filtering with a local tensor representation of curvilinear structures in 2D and 3D images. The proposed approach is validated on synthetic and real data, and is also compared to the state-of-the-art approaches. Our results show that the proposed approach achieves high-quality curvilinear structure enhancement in synthetic examples and in a wide range of 2D and 3D images. 
\end{abstract}
	\section{Introduction}\label{sec:intro}
The enhancement and detection of curvilinear structures are important and essential tasks in biomedical image processing. 
There is a wide range of curvilinear structure in biomedical imaging data, such as blood vessels, neurons, leaf veins, and fungal networks. 
Curvilinear structure enhancement is an important step, especially where the subjective quality of images of curvilinear structures is necessary for human interpretation. 

A wide range of curvilinear structure enhancement approaches have used mathematical morphology operations to enhance curvilinear structures in 2D and 3D images. 
The top-hat transform~\cite{haralick1987image} is a popular approach, which extracts bright features from a dark background that match the shape and orientation of a specified structuring element~\cite{zana2001segmentation}. 
This approach has been used to extract curvilinear structures in retinal~\cite{liao2014retinal} and fingerprint~\cite{bibiloni2017general} images. 

A local tensor representation~\cite{Knutsson1989} of an image measures how image structures change across dominant directions, and the eigenvalues and eigenvectors of the tensor can provide information that can be used to enhance, extract and analyse curvilinear structures.

In this paper, we combine these two approaches by representing curvilinear structures filtered by morphological operations in a local tensor representation of the image. 
We apply a multiscale top-hat with a line structuring element at different scales and orientations. 
Then, we produce a stack of top-hat images and combine them into a local tensor, find the eigenvalues to calculate vesselness and neuriteness to enhance the curvilinear structure in the biomedical images. This approach works with 2D and 3D images.
Compared with other existing approaches, the gathered results prove that our proposed approach achieves high-quality curvilinear structure enhancement in the synthetic examples and in a wide range of real 2D and 3D biomedical image types.
\section{Related Work} 
\label{RelatedWork}
Many curvilinear structure enhancement approaches for 2D and 3D images for a wide range of applications have been proposed in the literature to date. 
In this section, we list a small selection of the most relevant approaches divided into several subclasses according to the underlying concepts.

\subsection{Hessian Matrix-based Approaches}
A Hessian matrix-based image representation is constructed using responses of an image convolution with a set of matching filters, defined by second-order derivatives of the Gaussian at multiple scales~\cite{frangi1998multiscale,meijering2004design}. 
This concept is used to enhance and detect curve / tubular, sheet-like, and blob-like structures in the 2D and 3D images by exploring the relationships between eigenvectors and eigenvalues of the Hessian matrix. 
The three most common measurements proposed to date are: vesselness, neuriteness, and regularised volume ratio.

\subsubsection{Vesselness}
The vesselness measure~\cite{frangi1998multiscale} is calculated by computing the ratio of the eigenvalues of the Hessian matrix. The vesselness reaches its maximum when the scale and orientation of the filter matches the size and orientation of the local curvilinear structure. However, vesselness fails at junctions of curvilinear structures / networks due to the low filters responses.

\subsubsection{Neuriteness}
On the other hand, the neuriteness measure~\cite{meijering2004design} is based on a slightly modified Hessian matrix by adding a new tuning parameter. 
Neuriteness, in the same way as vesselness, fails at junctions of curvilinear structures / networks due to the low filters responses.s

\subsubsection{Volume Ratio-based Approach}
Hessian-based approaches rely on the eigenvalues and this leads to several problems: (1) eigenvalues are non-uniform throughout an elongated or rounded structure that has uniform intensity; (2) eigenvalues vary with image intensity; and (3) enhancement is not uniform across scales.
A recent volume ratio-based approach~\cite{jerman2016enhancement} aims to solve such problems by computing the ratio of Hessian eigenvalues to handle the low magnitudes of eigenvalues and uniform responses across different structures. 
This approach intends to intimate vascular elongated structures in 2D and 3D angiography images. 
However, it has drawbacks; despite enhancing the curvilinear structures, it also enhances the noise.
\subsection{Mathematical Morphology-based Approaches}
Morphological operations probe an image with a structuring element placed at all possible locations in the image and match it with the corresponding neighbourhood of pixels. 
This structuring element applied to an input image uses a set of operators (intersection, union, inclusion, complement). 
Morphological operations are easy to implement and are suitable for many shape-oriented problems. A great number of approaches have been proposed to enhance and detect the curvilinear structures based on different mathematical morphological transforms such as~\cite{soille2001directional,sazak2017multiscale2d,sazak2018multiscale3d}.
 
\subsubsection{Top-Hat Transform}
The top-hat transform has been widely used to enhance and detect curvilinear structures in retinal~\cite{zana2001segmentation} and aerial~\cite{roman2017top} images.
Zana and Klein~\cite{zana2001segmentation} enhance the curvilinear structures using the top-hat transform with line structuring elements at different directions and with a fixed scale. 
Then, they computed the sum of the top-hat along each direction, followed by a curvature measure that is calculated using a Laplacian of Gaussian. 
Thus, any small bright noise will be reduced and the contrast of curvilinear structures will be improved. 
%
%
%

\subsubsection{Path Operators Transform}
A mathematical morphology-based path opening and closing operation to detect the curvilinear structures in retinal images was introduced by~\cite{sigurdhsson2014automatic}. 
Recently, a new path operator called Ranking the Orientation Responses of Path Operators (RORPO) has been proposed to distinguish curvilinear objects from blob-like and planar structures in images~\cite{merveille20172d,merveille2018curvilinear}. 
The main disadvantage of the RORPO approach is its high computation cost when applied to large volume image datasets. 
Furthermore, this approach required an isotropic image resolution. 
\subsection{Phase Congruency Tensor-based Approaches}
Phase congruency (PC) was first introduced in~\cite{kovesi2003phase} and later combined with a local tensor to enhance curvilinear structures in 2D~\cite{obara2012contrast} and 3D~\cite{sazak2017contrast} images. 
The majority of Hessian-based approaches rely on image intensity, which leads to poor enhancement or suppression of finer and lower intensity vessels, where Phase Congruency Tensor-based approaches are image contrast-independent.
Moreover, the local tensor has a better representation of directions and the main advantage of using the local tensor is its ability to detect structures oriented in any direction. 
However, a major drawback of the PC-based concept is the complexity of its parameter space.
\subsection{Histogram-based Approaches}
Histogram-based approaches are the most popular technique for improving image contrast. 
Contrast Limited Adaptive Histogram Equalisation (CLAHE)~\cite{pisano1998contrast} is a widely used pre-processing stage in order to improve the local details of an image.
A major drawback of this method is its sensitivity to noise. 
An improvement proposed by~\cite{zhao2014retinal} employs the anisotropic diffusion filter to reduce the noise and smooth the image, especially near the boundary.
\subsection{Wavelet Transform-based Approaches}
The wavelet transform has been widely used for curvilinear structure enhancement in biomedical images. 
In~\cite{sihalath2010fingerprint}, the authors propose a new approach to enhance the curvilinear structures in fingerprint images by involving the second derivative of a Gaussian filter with a directional wavelet transform. 
Another approach combines the Discrete Wavelet Transform and morphological filter (opening and closing) to enhance curvilinear structures in MRI images~\cite{srivastava2011combination}. In addition, two different wavelets in parallel were applied in~\cite{demirel2011image} to achieve an enhanced  high-resolution image. 
In~\cite{bankhead2012fast}, the authors proposed an approach exploring the isotropic undecimated wavelet transform. 
However, similar to Hessian-based approaches, wavelet transform-based approaches fail to enhance low-intensity and fine curvilinear structures.
\subsection{Learning-based Approaches}
Recent learning-based methods are more suitable to deal with the scene complexity problem in natural images~\cite{lam2010general,azzopardi2015trainable,zhang2016robust}. In particular~\cite{annunziata2015scale}, proposed a new regression architecture based on the basis of filter banks learned by sparse convolutional coding to speed-up the training process. They are carefully designed hand-crafted filters (SCIRD-TS) which are modelling appearance properties of curvilinear structures. 
	\section{Method}\label{sec:method}
In this section, we introduce the proposed approach that consolidates the advantages of mathematical morphology and local tensor representation to enhance curvilinear structures in 2D/3D images.
Before explaining the proposed approach in detail, it is useful to provide some more background of the concepts that are applied in this paper.
\subsection{Background} 
\subsubsection{Mathematical Morphology and the Top-Hat Transform}
Mathematical morphology operations are a set of non-linear filtering methods, and almost all of them formed through a combination of two basic operators: dilation and erosion.

If $I(\mathbf{p})$ is a grey-scale image and $B(\mathbf{p})$ is structuring element where $\mathbf{p}$ denotes the pixel position $[x,y]^T$ in the 2D images and $[x,y,z]^T$ in the 3D images. Dilation, ($\oplus$) can be defined as the maximum of the points in a weighted neighbourhood determined by the structuring element, and mathematically:
\begin{equation}
(I \oplus B)(\mathbf{p}) = \sup_{\mathbf{x} \in E}[I(\mathbf{x})+B(\mathbf{p}-\mathbf{x})],
\end{equation}
where `$\sup$' is the supremum and $\mathbf{x} \in E$ denotes all points in Euclidean space within the image. Likewise, we mathematically represent erosion ($\ominus$), as the minimum of the points in the neighbourhood determined by the structuring element:
\begin{equation}
(I \ominus B)(\mathbf{p}) = \inf_{\mathbf{x} \in E}[I(\mathbf{x})+B(\mathbf{p}-\mathbf{x})],
\end{equation}
where `$\inf$' is the infimum. The behaviour of dilation is expanding bright areas and reducing dark areas, while erosion is expanding dark regions reducing bright areas~\cite{haralick1987image}.
From these two operators we can define two further commonly used morphological filters:
\begin{align}
	\text{opening}:    & \quad (I \circ B)(\mathbf{p}) = ((I \ominus B) \oplus B)(\mathbf{p})   \\
	\text{closing}:    & \quad (I \bullet B)(\mathbf{p}) = ((I \oplus B) \ominus B)(\mathbf{p})
\end{align}
where an opening ($\circ$) will preserve dark features and patterns, suppressing bright features, and a closing ($\bullet$) will preserve bright features whilst suppressing dark patterns.

By comparing the original image and the result of opening or closing, two region extraction operations, which are called top-hat (TH) and bottom-hat (BH) transform, and defined as follows;
\begin{align}
	\text{TH}= I(\mathbf{p}) - (I \circ B)(\mathbf{p})    \\
	\text{BT}= (I \bullet B)(\mathbf{p})-I(\mathbf{p})
\end{align}
The TH is usually used to extract bright structures, while BT is used to extract dark structures. 
\subsubsection{Vesselness and Neuriteness Measurements}
\paragraph{2D Vesselness}\label{sec:2DVes}
One of the most popular Hessian-based approaches that used the eigenvalues of the Hessian to compute the likeliness of an image region to contain vessels or other image ridges~\cite{frangi1998multiscale}. It is computed based on the ratio of eigenvalues of the Hessian matrix as follows:
\begin{equation}
\resizebox{.8\linewidth}{!}{$
	V_\sigma =\begin{cases}
	0, & \lambda_2>0 \\
	\text{exp}\Bigg(-\dfrac{R_\beta^2}{2\beta^2}\Bigg)
	\Bigg( 1-\text{exp}\bigg(-\dfrac{S^2}{2c^2}\bigg) \Bigg), & \text{otherwise}
	\end{cases}$}
\label{2DVesselness}
\end{equation}
where:
\begin{align}
	\resizebox{.5\linewidth}{!}{$
		R_\beta=\lambda _{1}/\lambda_{2}, \quad 
		S = \sqrt{\lambda_1^2 + \lambda_2^2}.$}
	\nonumber
\end{align}
The $\lambda_{1}$, $\lambda_{2}$ are eigenvalues of the Hessian matrix, and $\lambda_{1} \geq \lambda_{2}$. Where $\beta$ and $c$ are positive real user-defined parameters.
If the magnitude of both eigenvalues is small, i.e. the local image structure is likely to be part of the background, then the vesselness measure will be small. If one eigenvalue is small and the other large then the local structure is likely to be curvilinear and the vesselness measure is large. In case both of the eigenvalues magnitudes are large, then the structure is likely to be a blob and the vesselness measure will again small.
\paragraph{3D Vesselness}\label{sec:3DVess}
A 3D vesselness measure~\cite{frangi1998multiscale} is extended on the basis of all eigenvalues of the 3D Hessian matrix. Then, the vesselness for the 3D images is computed as follows:
\begin{equation}
\resizebox{.9\hsize}{!}{$
	V_\sigma = \begin{cases}
	0, & \lambda_2, \lambda_3>0 \\
	\text{exp}\Bigg(-\dfrac{R_{\beta}^2}{2\beta^2}\Bigg)
	\Bigg(1-\text{exp}\bigg(-\dfrac{R_\alpha^2}{2\alpha^2}\bigg)\Bigg) 
	\Bigg(1-\text{exp}\bigg(-\dfrac{S^2}{2c^2}\bigg)\Bigg), & \text{otherwise}	\end{cases}
	$}
\label{3Dvesselness}
\end{equation}	
and where;
\begin{align}
	\resizebox{.8\hsize}{!}{$
		S=\sqrt {\lambda_{1}^{ 2 }+\lambda _{ 2 }^{ 2 }+\lambda _{ 3 }^{ 2 } },\quad
		R_{\beta}=\lambda_{1}/{\sqrt{\lambda_2\lambda_3}},\quad
		R_{\alpha}=\lambda _{ 2 } / \lambda _{ 3 } .$}
	\nonumber
\end{align}

Similar to vesselness measure in 2D, the $\alpha $, $\beta$ and $c$ are real-valued positive user-defined parameters.
\paragraph{2D Neuriteness}  \label{sec:2DNeurite}
This method introduced by~\cite{meijering2004design} and designed to enhance low contrast and highly inhomogeneous neurites in the biomedical images. They changed the Hessian matrix by including a tuning parameter, alpha and derive two tuned eigenvalues $\lambda_1^{'}$ and $\lambda_2^{'}$ as follows: 
\begin{align} 
	\lambda_1^{'} =\lambda _ 1 +\alpha \lambda _{ 2 }, \nonumber\\ 
	\lambda_2^{'}=\lambda _{ 2 } +\alpha \lambda _{ 1 }.
\end{align}
Then, they consider the maximum and minimum eigenvalues across the whole image as describe below, and define a new neurite-enhancing metric $N_{\sigma}$.
\begin{align}
	\lambda_{max} =\max(|{\lambda_{1}}'|,|{\lambda_{2}}'|), \nonumber 
	\\ \lambda_{min} =  \min(\lambda_{max}).
	\label{lamdamax}
\end{align}
The neuritenees measurement define as:
\begin{equation}
N_\sigma=\begin{cases}
\dfrac{\lambda _{ max }}{\lambda _{ min }} & $if $ \lambda_{ max } <0 \\ 
0 &  $if $ \lambda _{ max }\ge 0 \end{cases},
\newcounter{mytempeqncnt}
\label{2DNeuriteness}
\end{equation}
where  $ { \lambda  }_{ i }^{ ' }$ are symbolised the normalized eigenvalues of modify Hessian matrix. The $\lambda_\text{min}$ denotes the smallest eigenvalue while $\lambda_\text{max}$ represents the largest one of eigenvalues. Additionally, line like structures which is dark $\left( \lambda _{ max } \ge 0 \right)$ are ignored  by the detector.
\paragraph{3D Neuriteness} \label{sec:3DNeurite}
The neuritenees measurement for the 3D image~\cite{meijering2004design} can define using a 3D modified Hessian matrix. Then, the 3D neuriteness measurement can define as:
\begin{equation}
N_\sigma=\begin{cases}
\dfrac{\lambda _{ max } }{\lambda _{ min }}    & \text{if } \lambda _{ max }<0 \\
0  &  \text{if }  \lambda _{ max } 	\geq 0
\end{cases},
\label{3DNeuriteness}
\end{equation}
and where; 
\begin{align}
	&	\lambda_1^{'} =\lambda _ 1 +\alpha \lambda _{ 2 }+\alpha \lambda _{ 3 },\nonumber\\ 
	&	\lambda_2^{'}=\lambda _{ 2 } +\alpha \lambda _{ 1 }+\alpha \lambda _{ 3 },\nonumber\\
	&	\lambda_3^{'}=\lambda _{ 3 } +\alpha \lambda _{ 1 }+\alpha \lambda _{ 2 },\nonumber\\
	&	\lambda_{max} =\max(|{\lambda_{1}}'|,|{\lambda_{2}}'|,|{\lambda_{3}}'),\nonumber\\
	&	\lambda_{min} =  \min(\lambda_{max}).\nonumber \notag
\end{align}
\subsection{Proposed Method Framework}
Since curvilinear structures can appear at different scales and directions in images, a top-hat transform using  multiscale and multi-directional structuring elements should be applied to detect them. 

The image is processed by using line structuring elements of different sizes (scale) and directions (orientations) and is then represented as a tensor, the Multiscale Top-Hat Tensor (MTHT), which intrinsically contains information on scale and orientation.
Then, through the use of its eigenvalues and eigenvectors, vesselness and neuriteness are calculated to enhance curvilinear structures. The details of the proposed approach are given below.

\subsubsection{Multiscale Top-Hat Transform}
For a given 2D/3D grayscale image $I(\mathbf{p})$, where $\mathbf{p}$ donates the pixel position, a stack of 2D/3D line structuring elements $B_{{\sigma}_i,\mathbf{u}_j}$, for $m$ different scales $\sigma_i$ and $n$ different orientations $\mathbf{u}_j$, is defined.

In 2D, the $\mathbf{u}_j$ orientation of line structuring element is defined as follows:
\begin{equation}
\label{eq:unitvector}
\mathbf{u}_j = \left[cos({\theta}_j),sin({\theta}_j)\right]^{T},
\end{equation}
where ${\theta}_j \in [0; 180)$. 

In 3D, as proposed in~\cite{hacihaliloglu20062a,sazak2017contrast}, a point distribution on the sphere of unit radius is used to define the orientation $\mathbf{u}_j$ of the 3D line structuring element as follows:

\begin{equation}
\mathbf{u}_j = [sin(\theta_j)cos(\phi_j), sin(\theta_j)sin(\phi_j), cos(\theta_j)]^T,
\end{equation}
where ${\theta}_j \in [0; 180]$ and  $\phi_j \in [0; 360)$.

Then, we produced a top-hat image using a line structuring element defined by scale $\sigma_i$ and orientation $\mathbf{u}_j$ as follows:
\begin{equation}
TH(\mathbf{p})_{{\sigma}_i,\mathbf{u}_j}=I(\mathbf{p}) - (I \circ B_{\sigma_i,\mathbf{u}_j})(\mathbf{p}).
\label{TH}
\end{equation}

\subsubsection{Tensor Representation}
In general, the tensor representation of an image can provide information about how much the image differs along and across the dominant orientations within a particular region~\cite{Knutsson1989}. 

In our case, the local tensor $T(\mathbf{p})_{\sigma_i}$ representation of an image $I(\mathbf{p})$ is generated by combining the bank of top-hat images from Equation~\ref{TH} as follows:
\begin{equation}
T(\mathbf{p})_{\sigma_i}=\sum_{j=1}^{n}\|TH(\mathbf{p})_{\sigma_i,\mathbf{u}_j}\|(\mathbf{u}_j\mathbf{u}_j^T).
\label{eq:tensor}
\end{equation}
\subsubsection{MTHT Vesselness}
As described in Section~\ref{RelatedWork}, piecewise curvilinear segments can be detected by analysing the relations between eigenvalues and eigenvectors of the locally calculated Hessian~\cite{frangi1998multiscale}. 
In a similar way~\cite{obara2012contrast}, the vesselness of the proposed approach is defined where the eigenvalues of the Hessian matrix are substituted with those of the MTHT. 
Finally, multiscale vesselness, for a given set of $m$ scales can be calculated as follows:
\begin{align}
	V = \underset{{i}} \max \left(V_{\sigma_i}\right).
\end{align}
\subsubsection{MTHT Neuriteness}
When combining the neuriteness with our approach, it is necessary to modify the neuriteness measurement introduced by ~\cite{meijering2004design} for 2D and 3D images respectively.
In~\cite{meijering2004design}, they normalised eigenvalues correspondingly to the smaller absolute eigenvalue which is a negative value. 
Whereas, in our approach, we used a morphological line structuring element instead of the second order derivative of the Gaussian function used by~\cite{meijering2004design}, so the smaller absolute eigenvalue will be equal to $0$. 
The modify neuriteness equation is:
\begin{equation}
N_{\sigma_i}=\begin{cases}
\dfrac{\lambda}{\lambda _{ max }} & $if $ \lambda>0 \\ 
0 &  $if $ \lambda = 0 \end{cases},
\label{2DNeuritenessModify}
\end{equation}
where $\lambda$ is the larger in the magnitude of the two eigenvalues $\lambda_{1}$ and $\lambda_{2}$ for 2D images or the larger in the magnitude of the three eigenvalues  $\lambda_{1}$,  $\lambda_{2}$ and $\lambda_{3}$ for 3D images. $\lambda_{max}$ denotes the largest $\lambda$ over all pixels in the image. Similar to vesselness, a multiscale neuriteness can be calculated as:
\begin{align}
	N = \underset{{i}} \max \left(N_{\sigma_i}\right).
\end{align}
	\section{Results} \label{sec:result}
In this section, we present quantitative and qualitative validations for the proposed approach against both synthetic and real-world 2D and 3D imaging data. We then compare the results with state-of-the-art approaches. 
In order to validate the approach quantitatively in 2D and 3D images, we calculate the Receiver Operating Characteristic (ROC) curve and the Area Under the Curve (AUC), further details can be found in~\cite{fawcett2006introduction}.

\renewcommand{\cWidth}{0.087}
\newcommand{\ch}{1.2 cm}
\begin{figure*}[h!]
	\centering
	\begin{subfigure}[t]{\cWidth\linewidth}\includegraphics[width=\linewidth]{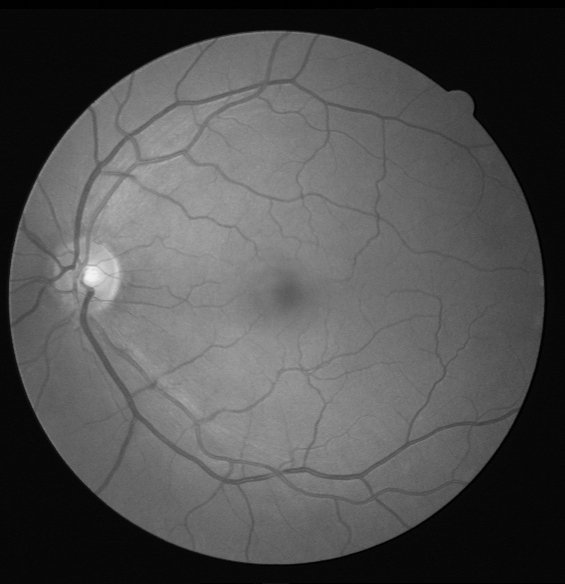}
	\end{subfigure}
	\begin{subfigure}[t]{\cWidth\linewidth}\includegraphics[width=\linewidth]{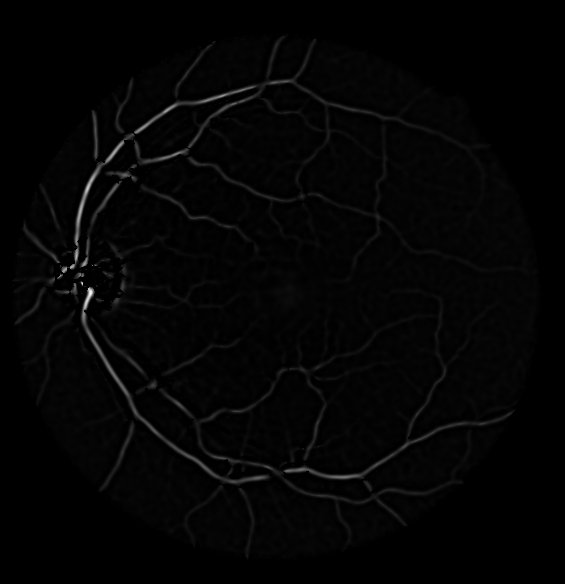}
	\end{subfigure}
\begin{subfigure}[t]{\cWidth\linewidth}\includegraphics[width=\linewidth]{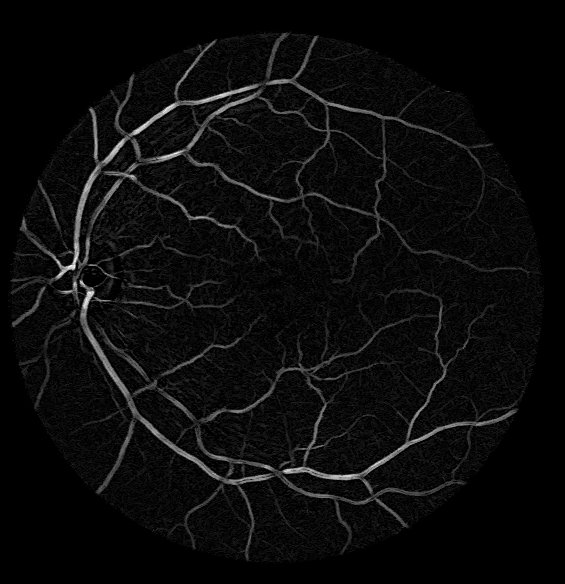}
\end{subfigure}
	\begin{subfigure}[t]{\cWidth\linewidth}\includegraphics[width=\linewidth]{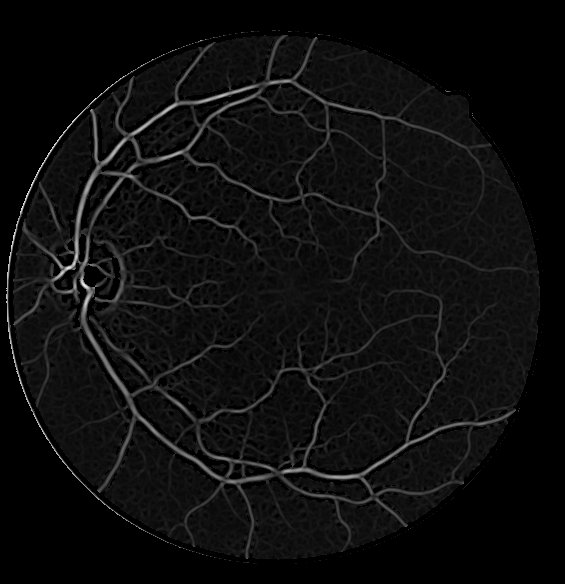}
	\end{subfigure}
	\begin{subfigure}[t]{\cWidth\linewidth}\includegraphics[width=\linewidth]{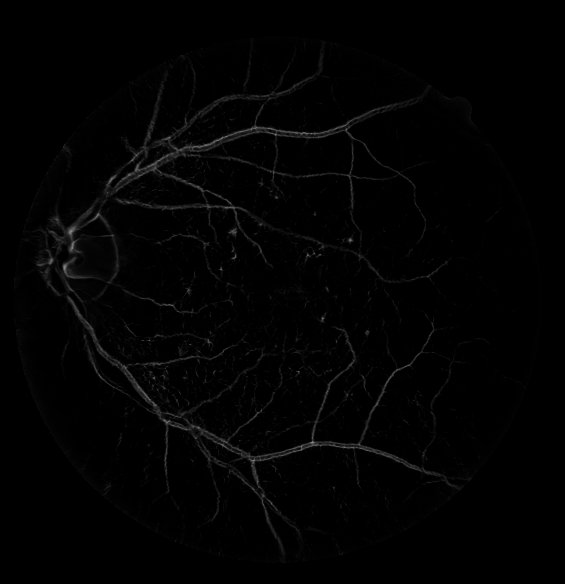}
\end{subfigure}
\begin{subfigure}[t]{\cWidth\linewidth}\includegraphics[width=\linewidth]{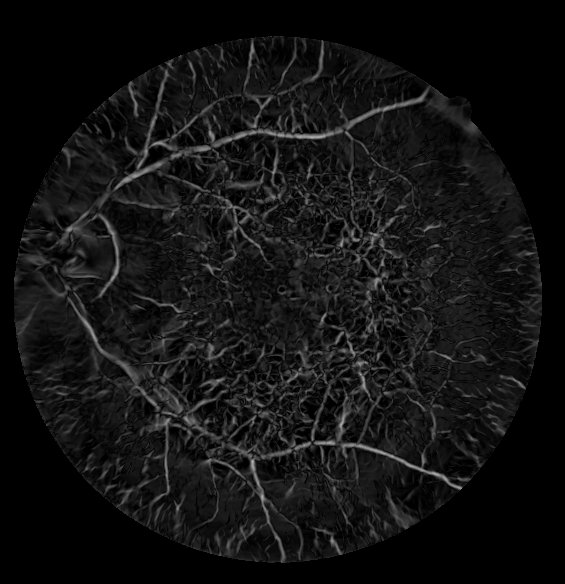}
\end{subfigure}
	\begin{subfigure}[t]{\cWidth\linewidth}\includegraphics[width=\linewidth]{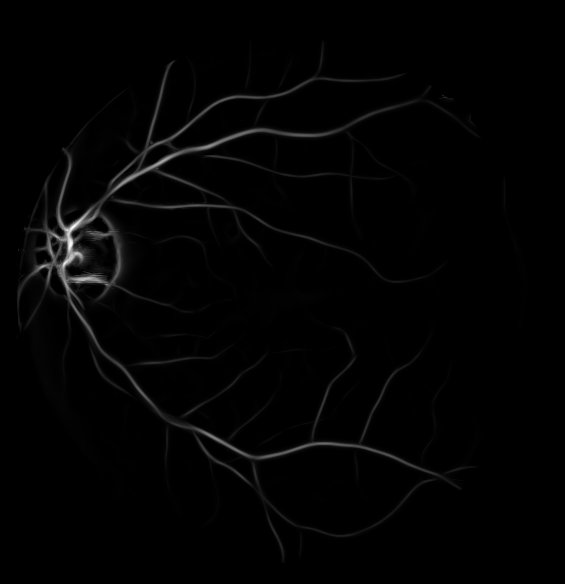}
\end{subfigure}
	\begin{subfigure}[t]{\cWidth\linewidth}\includegraphics[width=\linewidth]{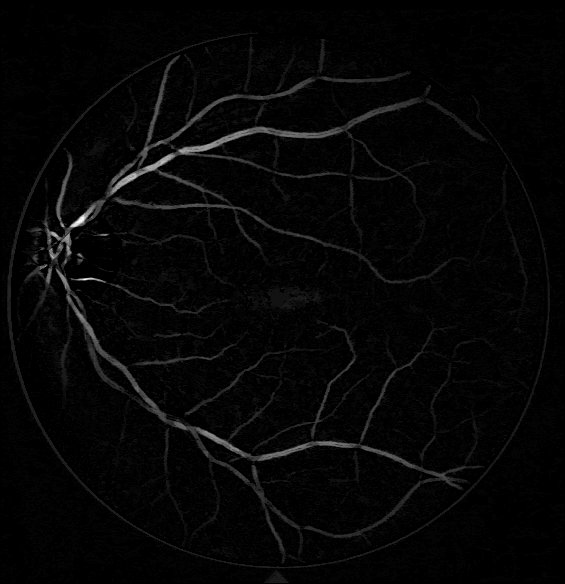}
\end{subfigure}
	\begin{subfigure}[t]{\cWidth\linewidth}\includegraphics[width=\linewidth]{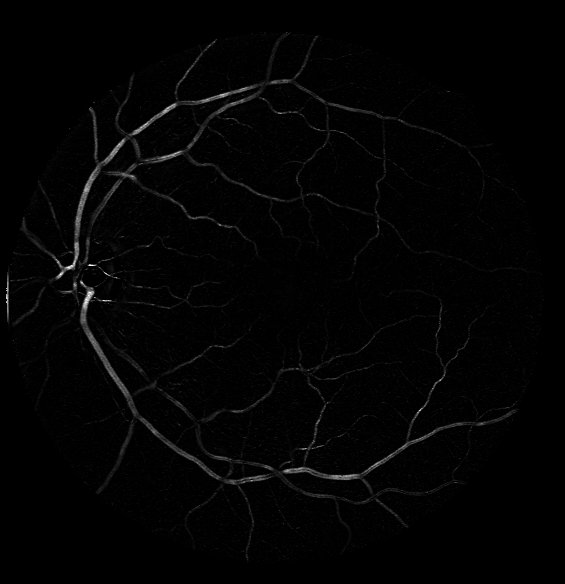}
	\end{subfigure}
    \begin{subfigure}[t]{\cWidth\linewidth}\includegraphics[width=\linewidth]{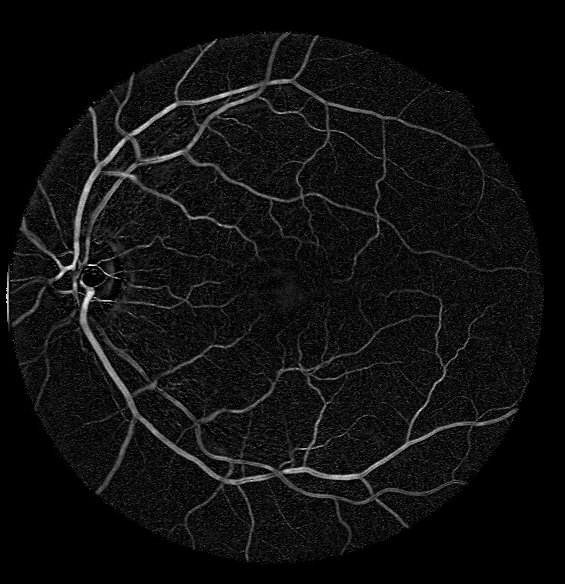}
    \end{subfigure}\\ \vskip4pt
	\begin{subfigure}[t]{\cWidth\linewidth}\includegraphics[width=\linewidth]{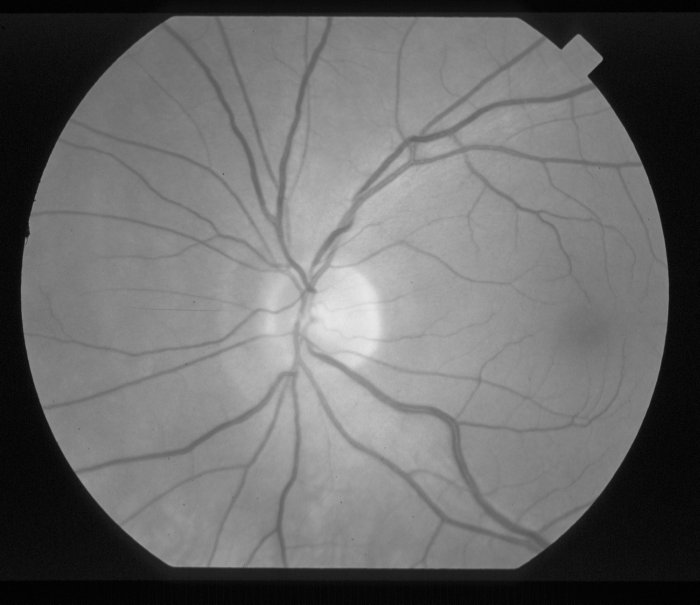}
	\end{subfigure}
	\begin{subfigure}[t]{\cWidth\linewidth}\includegraphics[width=\linewidth]{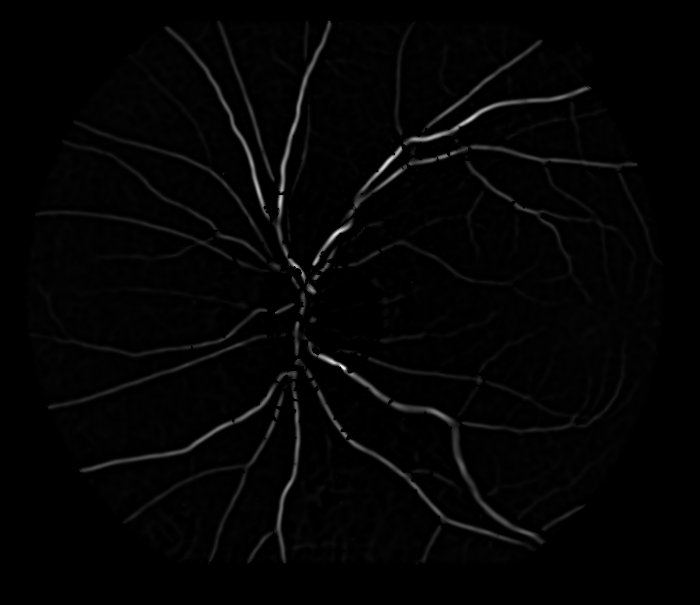}
	\end{subfigure}
    \begin{subfigure}[t]{\cWidth\linewidth}\includegraphics[width=\linewidth]{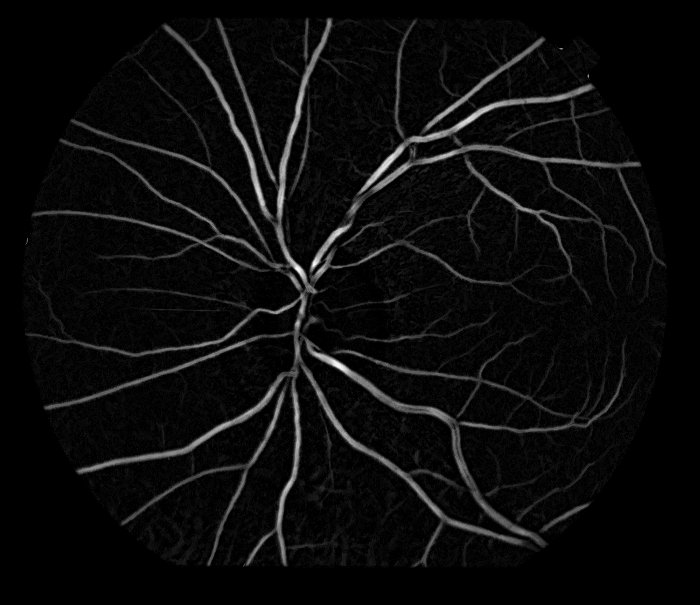}
    \end{subfigure}
	\begin{subfigure}[t]{\cWidth\linewidth}\includegraphics[width=\linewidth]{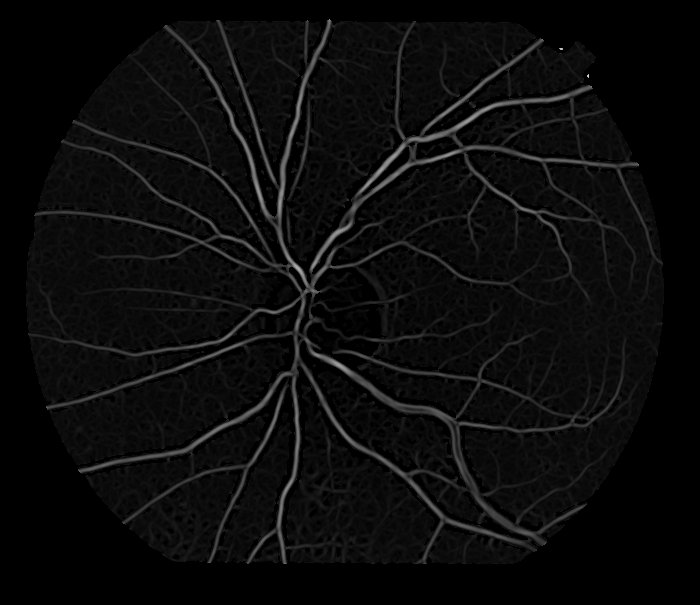}
	\end{subfigure}
    \begin{subfigure}[t]{\cWidth\linewidth}\includegraphics[width=\linewidth]{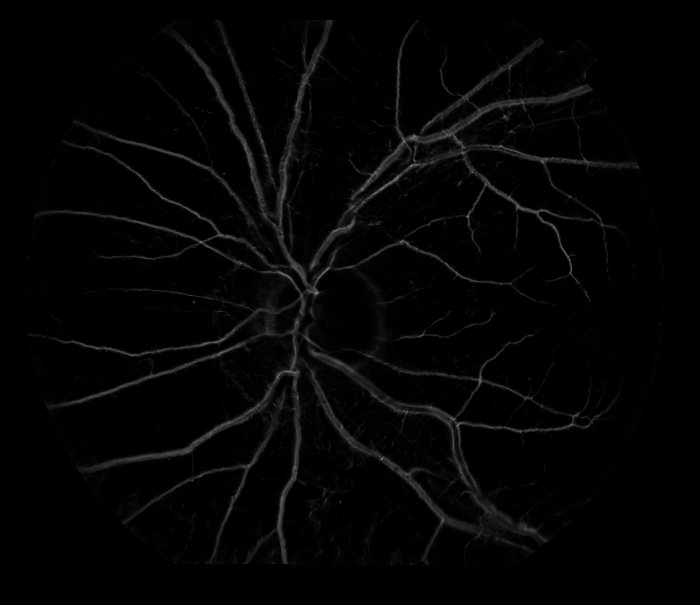}
    \end{subfigure}
    \begin{subfigure}[t]{\cWidth\linewidth}\includegraphics[width=\linewidth]{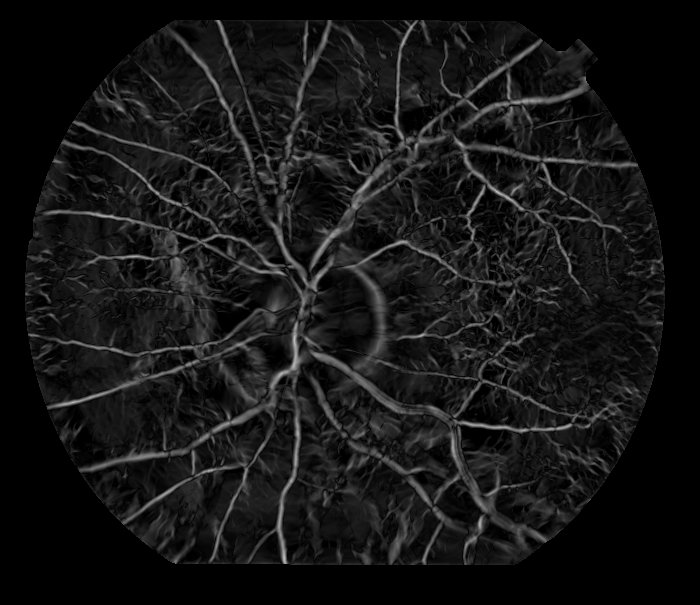}
    \end{subfigure}
    \begin{subfigure}[t]{\cWidth\linewidth}\includegraphics[width=\linewidth]{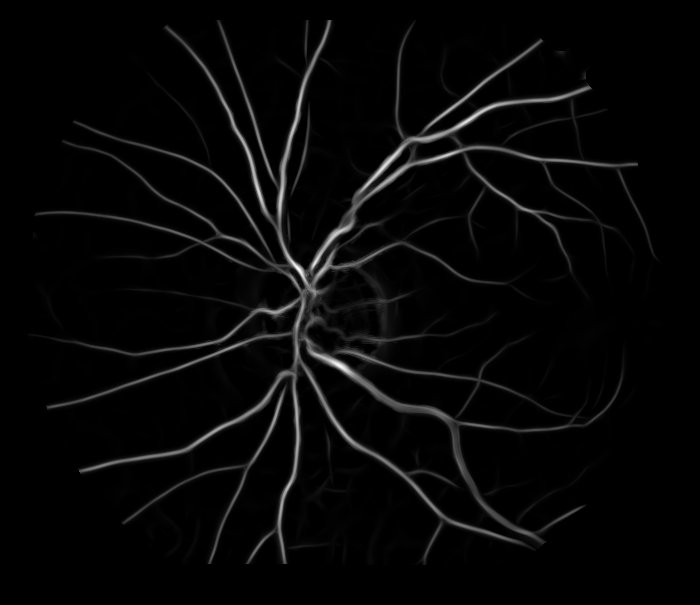}
    \end{subfigure}
    \begin{subfigure}[t]{\cWidth\linewidth}\includegraphics[width=\linewidth]{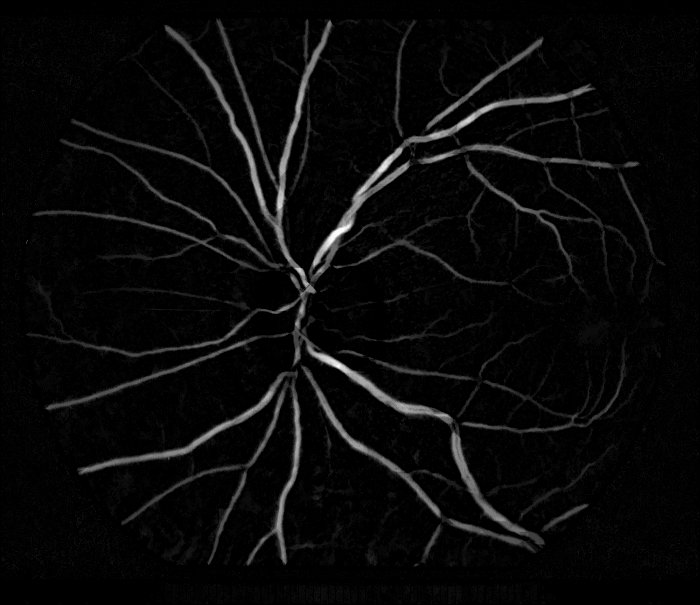}
    \end{subfigure}
	\begin{subfigure}[t]{\cWidth\linewidth}\includegraphics[width=\linewidth]{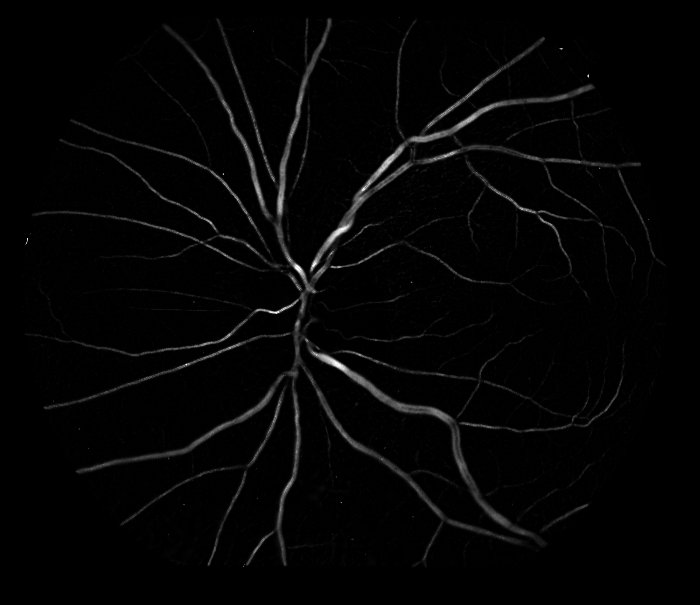}
	\end{subfigure}
	\begin{subfigure}[t]{\cWidth\linewidth}\includegraphics[width=\linewidth]{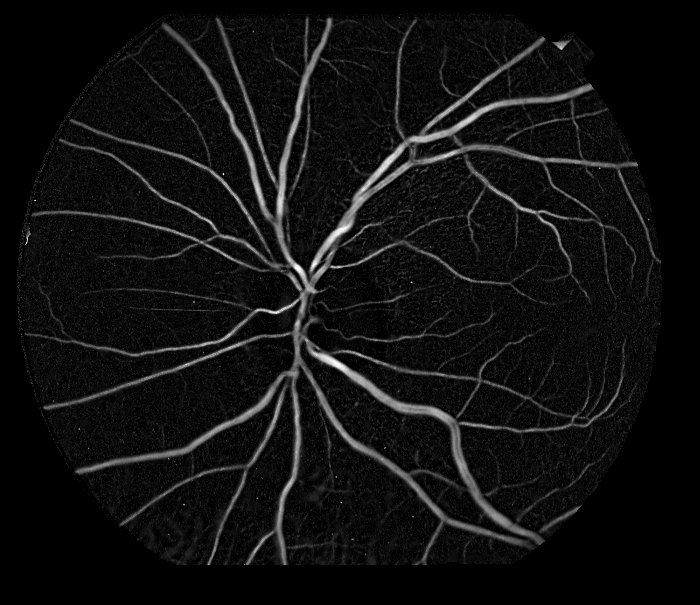}
	\end{subfigure}\\ \vskip4pt
	\begin{subfigure}[t]{\cWidth\linewidth}\includegraphics[width=\linewidth,height=\ch]{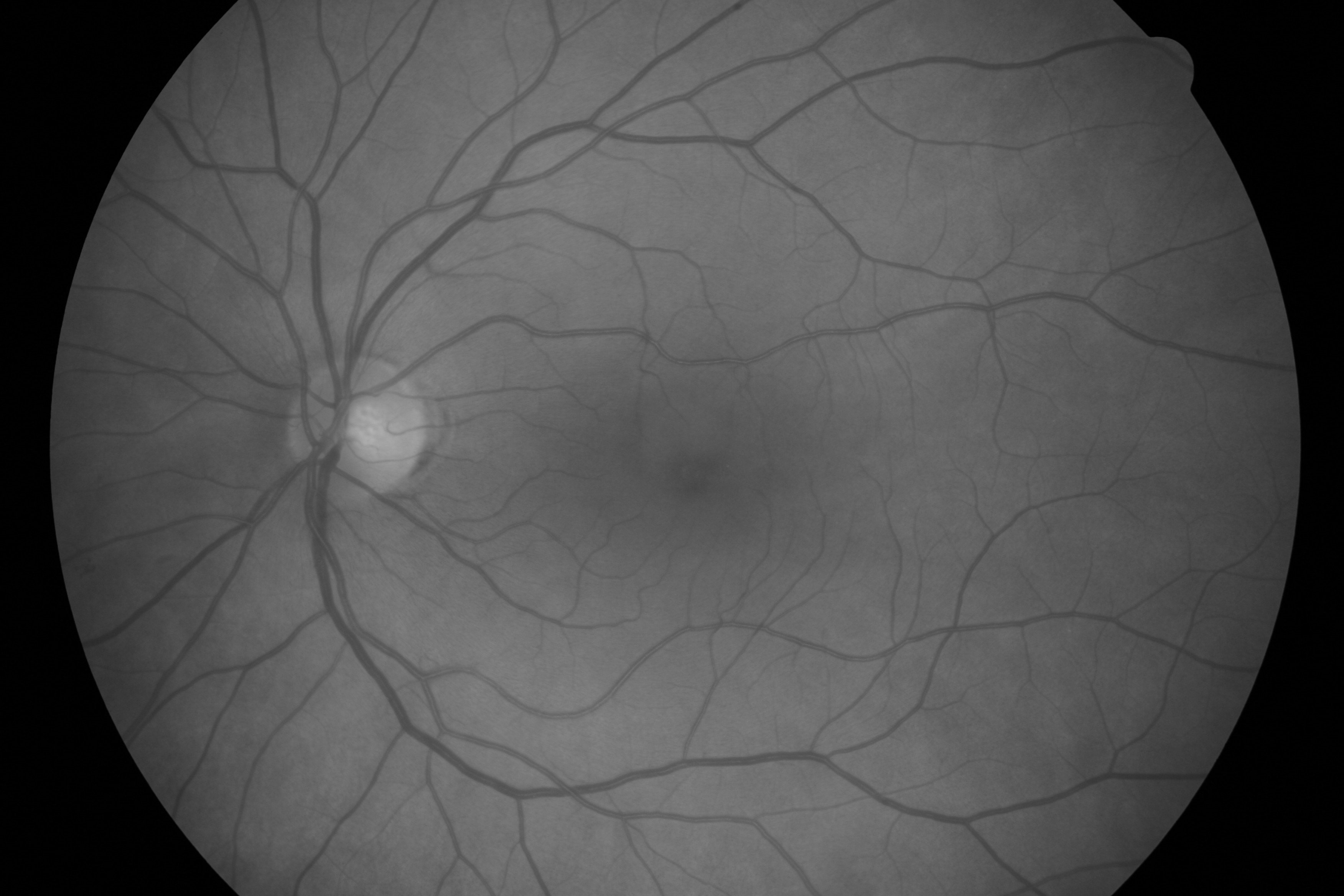}
	\caption{\quad}
	\end{subfigure}
	\begin{subfigure}[t]{\cWidth\linewidth}\includegraphics[width=\linewidth,height=\ch]{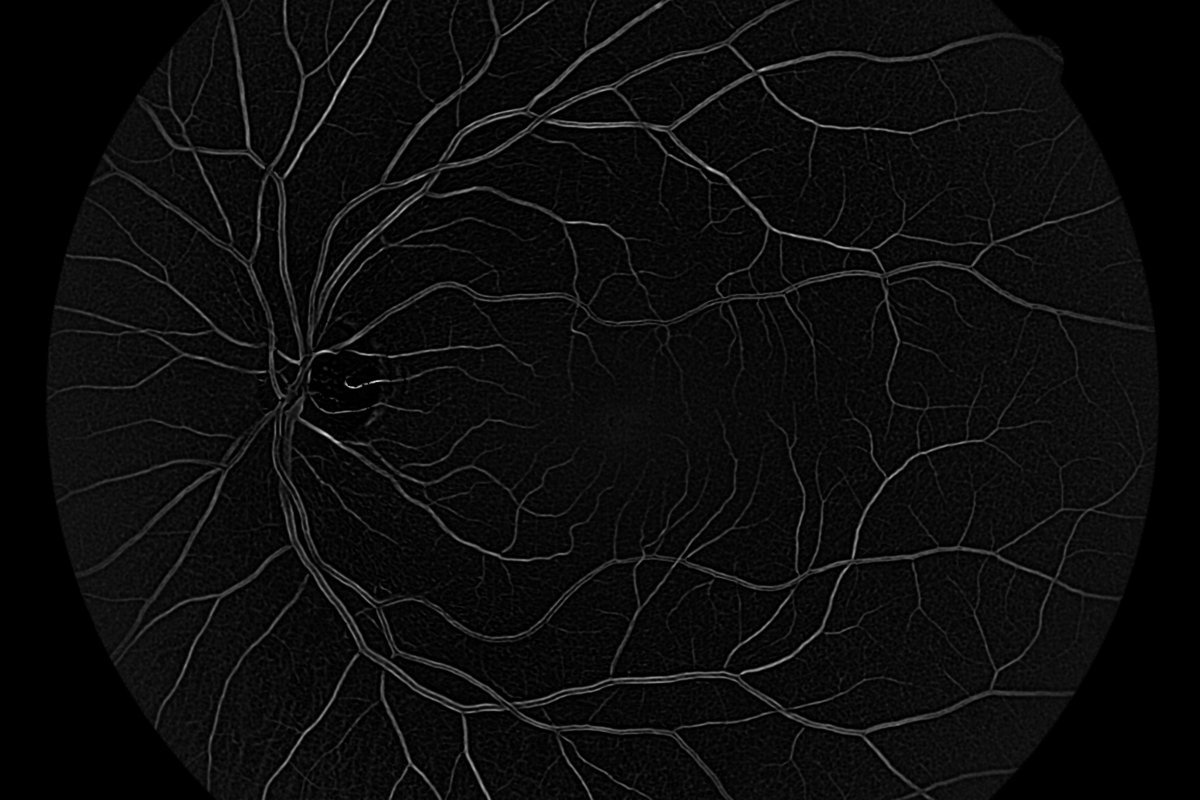}
		\caption{\quad}
	\end{subfigure}
    \begin{subfigure}[t]{\cWidth\linewidth}\includegraphics[width=\linewidth,height=\ch]{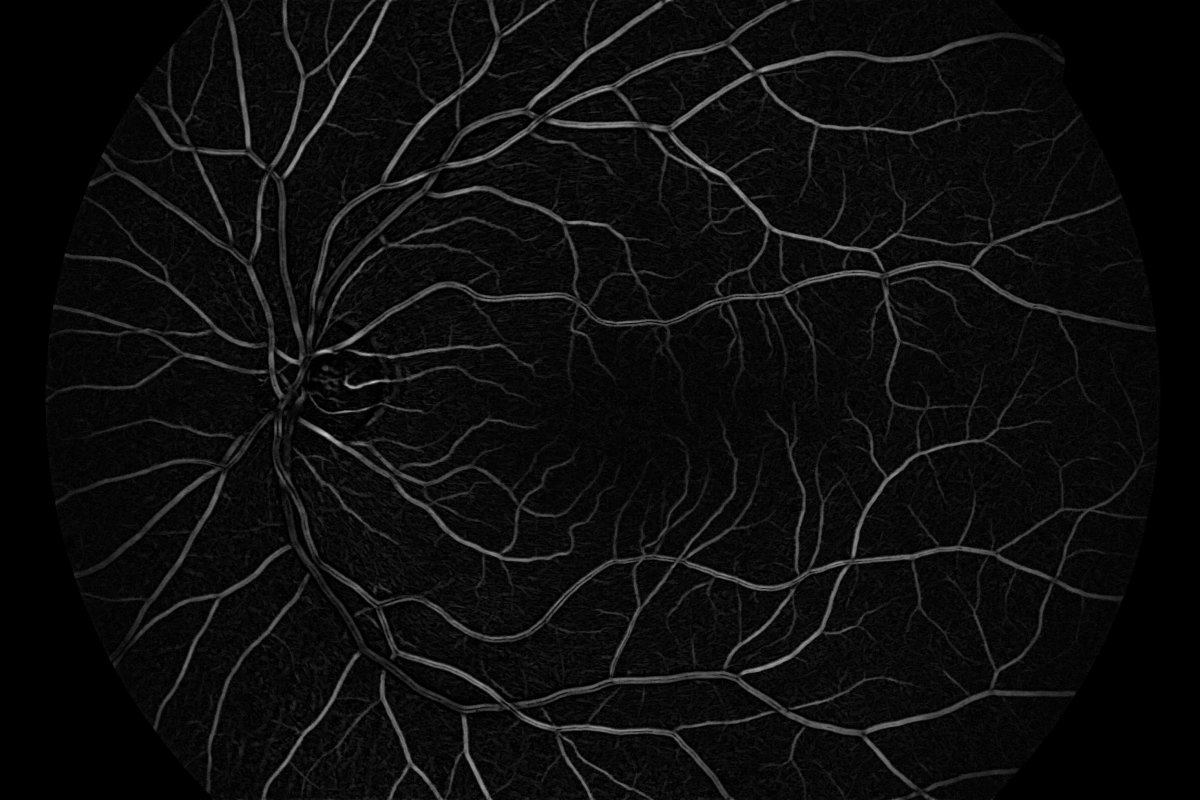}
	\caption{}
    \end{subfigure}
	\begin{subfigure}[t]{\cWidth\linewidth}\includegraphics[width=\linewidth,height=\ch]{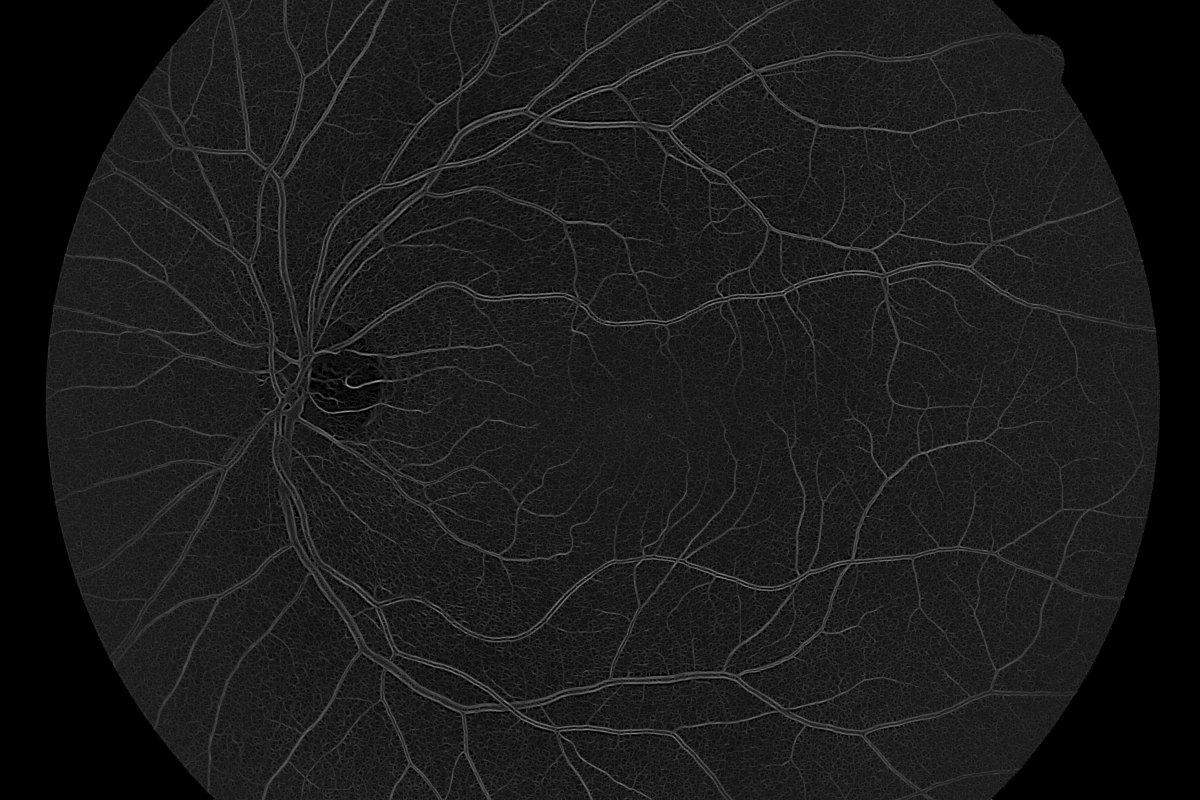}
		\caption{\quad}
	\end{subfigure}
	\begin{subfigure}[t]{\cWidth\linewidth}\includegraphics[width=\linewidth,,height=\ch]{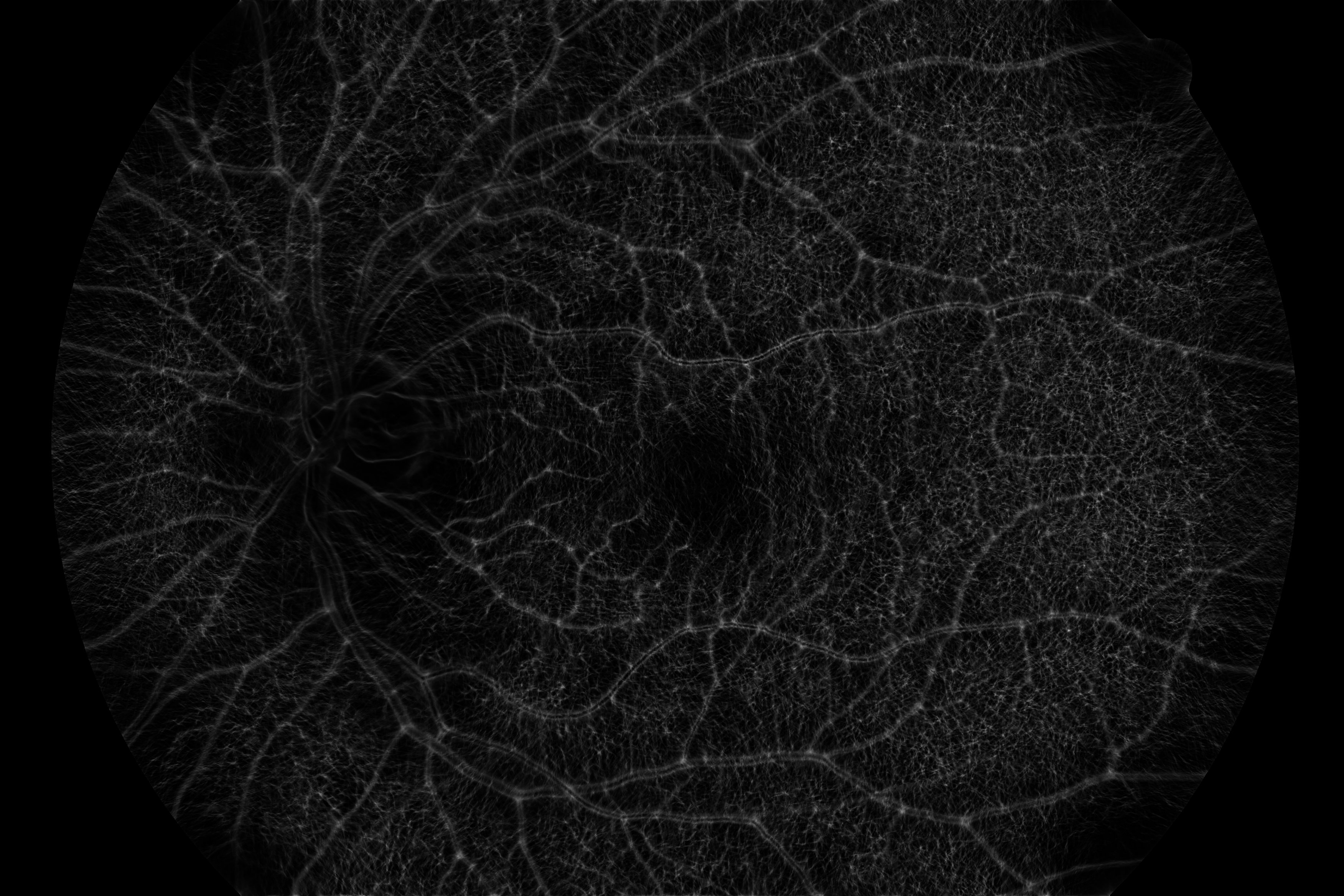}
	\caption{\quad}
\end{subfigure}
\begin{subfigure}[t]{\cWidth\linewidth}\includegraphics[width=\linewidth,,height=\ch]{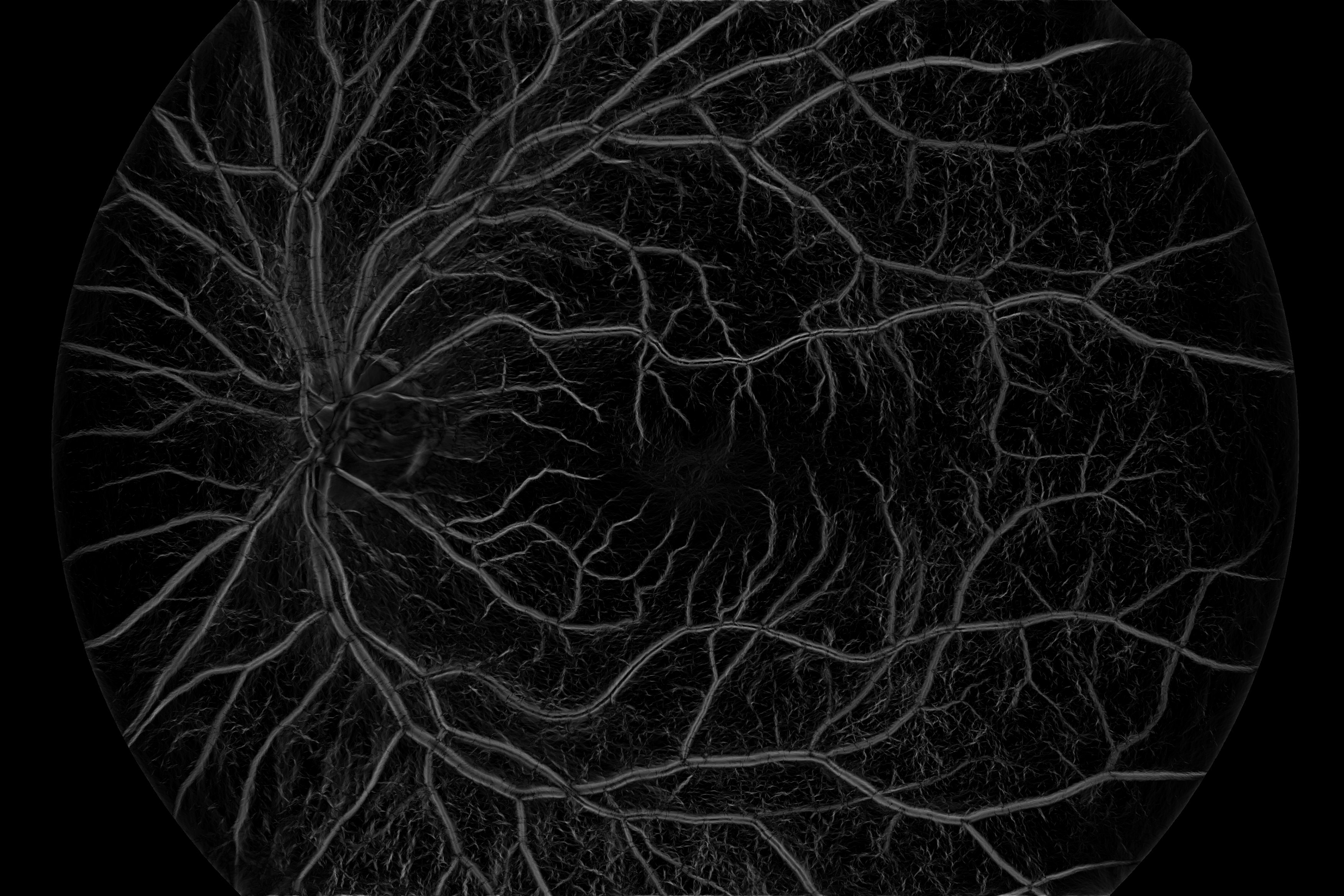}
	\caption{\quad}
\end{subfigure}
   \begin{subfigure}[t]{\cWidth\linewidth}\includegraphics[width=\linewidth,,height=\ch]{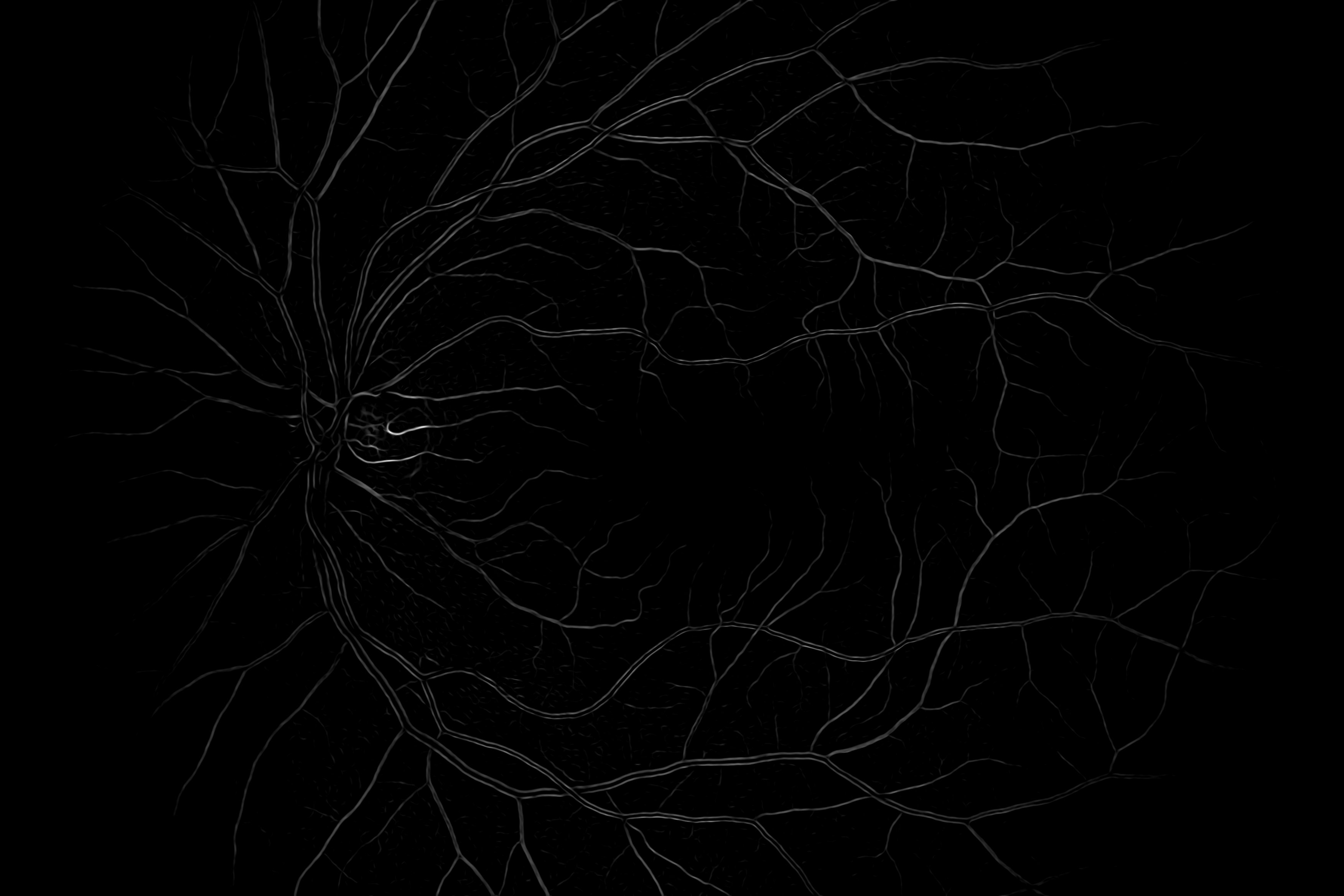}
	\caption{\quad}
   \end{subfigure}
    \begin{subfigure}[t]{\cWidth\linewidth}\includegraphics[width=\linewidth,height=\ch]{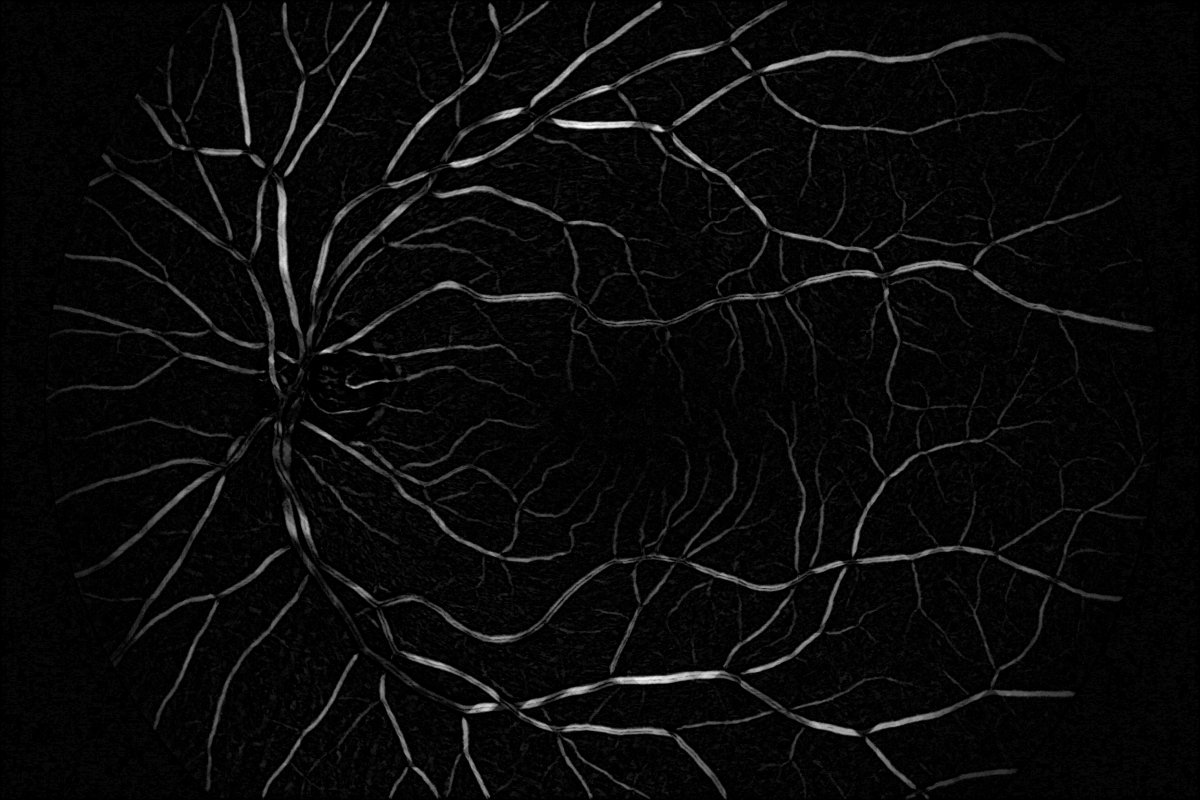}
	\caption{\quad}
    \end{subfigure}
	\begin{subfigure}[t]{\cWidth\linewidth}\includegraphics[width=\linewidth,height=\ch]{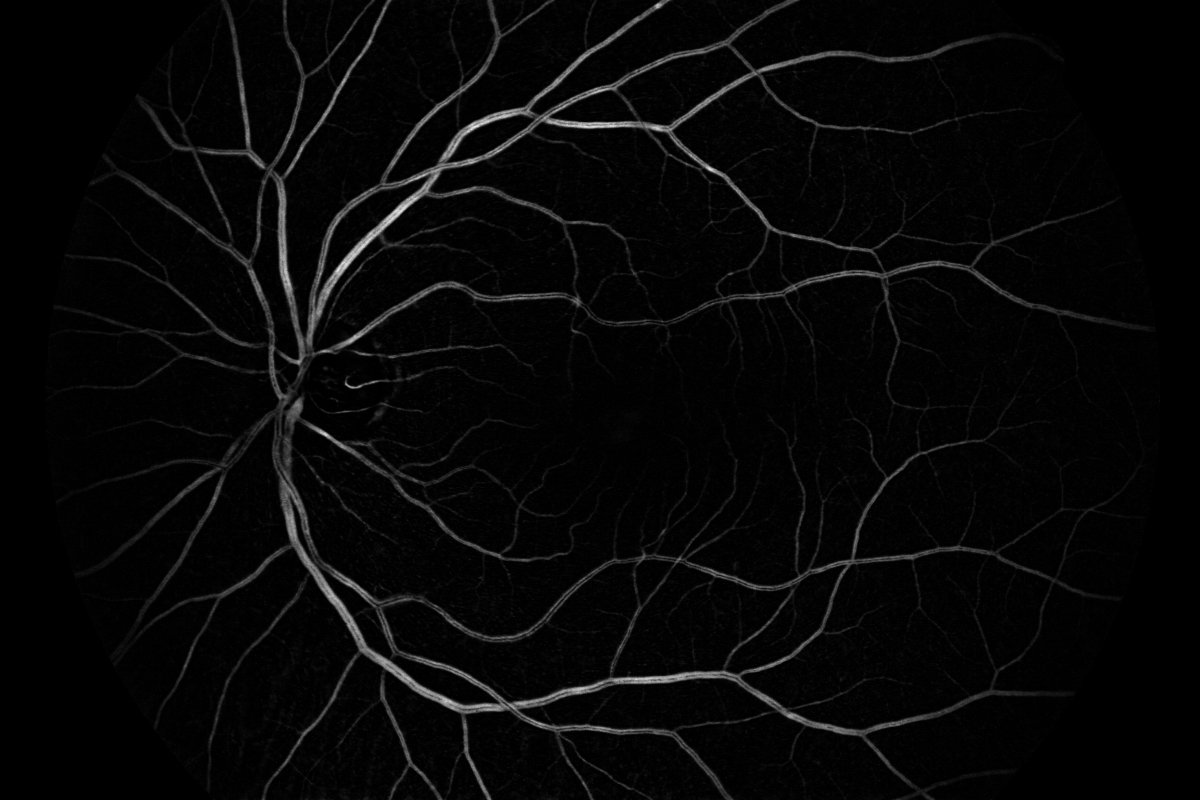}
	\caption{\quad}
	\end{subfigure}
	\begin{subfigure}[t]{\cWidth\linewidth}\includegraphics[width=\linewidth,height=\ch]{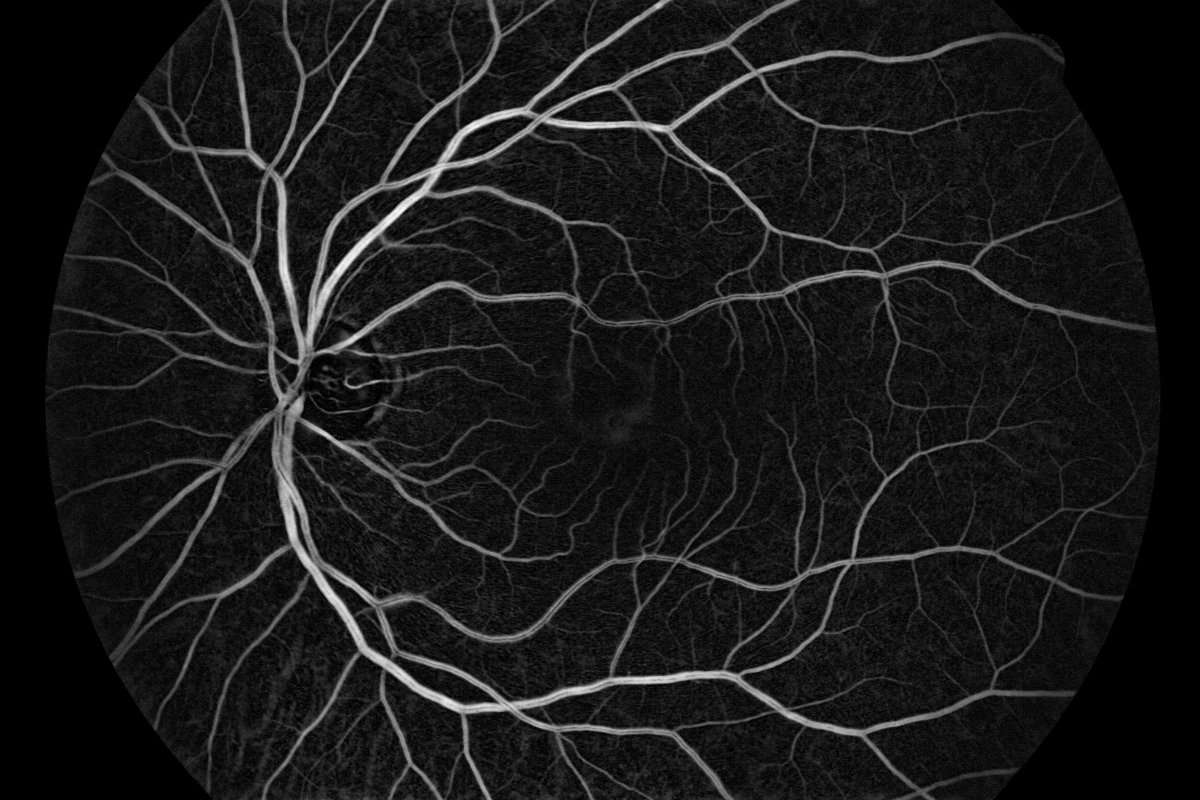}
		\caption{\quad}
	\end{subfigure}
	\caption{A selection of 2D retina images alongside the enhanced images from the state-of-the-art approaches. (a) is the original grayscale images from the DRIVE, STARE, and HRF retina dataset, (b) Vesselness~\cite{frangi1998multiscale}, (c) Zana's top-hat~\cite{zana2001segmentation}, (d) Neuriteness~\cite{meijering2004design}, (e) PCT vesselness~\cite{obara2012contrast}, (f) PCT neuriteness~\cite{obara2012contrast}, (g) SCIRD-TS~\cite{annunziata2015scale}, (h) RORPO~\cite{merveille20172d}, (i) MTHT vesselness, and (j) MTHT neuriteness.}
	\label{fig:rtina}
\end{figure*}
\newcommand{\sfigure}{0.3}
\begin{figure*}[t!]
	\begin{subfigure}{\sfigure\linewidth}
		\begin{tikzpicture}
		\begin{axis}[
        width =1.1\linewidth,
		xlabel={False Positive Rate (1-Specificity)},
		ylabel={True Positive Rate (1-Sensitivity)},
		ylabel near ticks,
		label style={font=\tiny},
		tick label style={font=\tiny},
		ytickmin=0, ymax=1.05,
		xtickmin=0, xtickmax=1.05,
		enlargelimits=false,
		grid=major, 
		grid style={dashed,gray!30}, 
		]
	\addplot[color=black,loosely dashed] table [x index=0, y index=1, col sep=comma] {images/datFiles/driveO.dat};
	\addplot[color=brewerDark1,line width=1pt] table [x index=0, y index=1, col sep=comma] {images/datFiles/driveV.dat};
	\addplot[color=brewerDark5,line width=1pt] table [x index=0, y index=1, col sep=comma] {images/datFiles/drivePZ.dat};
	\addplot[color=brewerDark2,line width=1pt] table [x index=0, y index=1, col sep=comma] {images/datFiles/driveN.dat};
	\addplot[color=brewerDark3,line width=1pt] table [x index=0, y index=1, col sep=comma] {images/datFiles/DRIVEPCTV.dat};
	\addplot[color=brewerDark4,line width=1pt] table [x index=0, y index=1, col sep=comma] {images/datFiles/DRIVEPCTN.dat};
	\addplot[color=brewerDark7,line width=1pt] table [x index=0, y index=1, col sep=comma] {images/datFiles/SCIRDdrive.dat};
	\addplot[color=brewerDark6,line width=1pt] table [x index=0, y index=1, col sep=comma] {images/datFiles/drivePO.dat};
	\addplot[color=red,line width=1pt] table [x index=0, y index=1, col sep=comma] {images/datFiles/drivePV.dat};
	\addplot[color=black,line width=1pt] table [x index=0, y index=1, col sep=comma] {images/datFiles/drivePN.dat};
\end{axis} 
\end{tikzpicture}
\caption{DRIVE}
\end{subfigure}\quad
	\begin{subfigure}{\sfigure\linewidth}
	\begin{tikzpicture}
	\begin{axis}[
    width =1.1\linewidth,
	xlabel={False Positive Rate (1-Specificity)},
	ylabel={True Positive Rate (1-Sensitivity)},
	ylabel near ticks,
	label style={font=\tiny},
	tick label style={font=\tiny},
	ytickmin=0, ymax=1.05,
	xtickmin=0, xtickmax=1.05,
	enlargelimits=false,
	grid=major, 
	grid style={dashed,gray!30}, 
	]
	\addplot[color=black,loosely dashed] table [x index=0, y index=1, col sep=comma] {images/datFiles/STAREO.dat};
	\addplot[color=brewerDark1,line width=1pt] table [x index=0, y index=1, col sep=comma] {images/datFiles/STAREV.dat};
	\addplot[color=brewerDark5,line width=1pt] table [x index=0, y index=1, col sep=comma] {images/datFiles/STAREZ.dat};
	\addplot[color=brewerDark2,line width=1pt] table [x index=0, y index=1, col sep=comma] {images/datFiles/STAREN.dat};
	\addplot[color=brewerDark3,line width=1pt] table [x index=0, y index=1, col sep=comma] {images/datFiles/STAREPCTV.dat};
	\addplot[color=brewerDark4,line width=1pt] table [x index=0, y index=1, col sep=comma] {images/datFiles/STAREPCTN.dat};
	\addplot[color=brewerDark7,line width=1pt] table [x index=0, y index=1, col sep=comma] {images/datFiles/SCIRDstare.dat};
	\addplot[color=brewerDark6,line width=1pt] table [x index=0, y index=1, col sep=comma] {images/datFiles/STAREPO.dat};
	\addplot[color=red,line width=1pt] table [x index=0, y index=1, col sep=comma] {images/datFiles/STAREPV.dat};
	\addplot[color=black,line width=1pt] table [x index=0, y index=1, col sep=comma] {images/datFiles/STAREPN.dat};
	\end{axis}
	\end{tikzpicture}
	\caption{STARE}
\end{subfigure}\quad
%
	\begin{subfigure}{\sfigure\linewidth}
	\begin{tikzpicture}
	\begin{axis}[
	width =1.1\linewidth,
	xlabel={False Positive Rate (1-Specificity)},
	ylabel={True Positive Rate (1-Sensitivity)},
	ylabel near ticks,
	label style={font=\tiny},
	tick label style={font=\tiny},
	ytickmin=0, ymax=1.05,
	xtickmin=0, xtickmax=1.05,
	enlargelimits=false,
	legend entries = {Raw image, Vesselness, Zana's top-hat, Neuriteness, PCT vesselness, PCT neuriteness, SCIRD-TS, RORPO, MTHT vesselness, MTHT neuritenss},
	legend columns = 5,
	legend style={font=\footnotesize,line width=2pt, draw=none,},
	legend to name=leg3,
	grid=major, 
	grid style={dashed,gray!30}, 
	]
	\addplot[color=black,loosely dashed] table [x index=0, y index=1, col sep=comma] {images/datFiles/STAREO.dat};
	\addplot[color=brewerDark1,line width=1pt] table [x index=0, y index=1, col sep=comma] {images/datFiles/STAREV.dat};
	\addplot[color=brewerDark5,line width=1pt] table [x index=0, y index=1, col sep=comma] {images/datFiles/HRFHZ.dat};
	\addplot[color=brewerDark2,line width=1pt] table [x index=0, y index=1, col sep=comma] {images/datFiles/HRFHN.dat};
	\addplot[color=brewerDark3,line width=1pt] table [x index=0, y index=1, col sep=comma] {images/datFiles/HRFHPCTV.dat};
	\addplot[color=brewerDark4,line width=1pt] table [x index=0, y index=1, col sep=comma] {images/datFiles/HRFHPCTN.dat};
	\addplot[color=brewerDark7,line width=1pt] table [x index=0, y index=1, col sep=comma] {images/datFiles/SCIRDhrf_healthy.dat};
	\addplot[color=brewerDark6,line width=1pt] table [x index=0, y index=1, col sep=comma] {images/datFiles/HarfPO.dat};
	\addplot[color=red,line width=1pt] table [x index=0, y index=1, col sep=comma] {images/datFiles/HarfPV.dat};
	\addplot[color=black,line width=1pt] table [x index=0, y index=1, col sep=comma] {images/datFiles/HarfPN.dat};
	\end{axis}
	\end{tikzpicture}
	\caption{HRF (healthy)}
\end{subfigure}
	\centering
	\pgfplotslegendfromname{leg3}	
\caption{Mean ROC curves are calculated for all the 2D retina images in: (a) DRIVE, (b) STARE, and (c) HRF datasets enhanced using the state-of-the-art approaches alongside the proposed MTHT Vesselness and MTHT Neuiteness (see legend for colours). Correspondingly, the mean AUC values for all datasets can be found in Table~\ref{tab:auc2}.}
\label{fig:retinaROC}
\end{figure*}
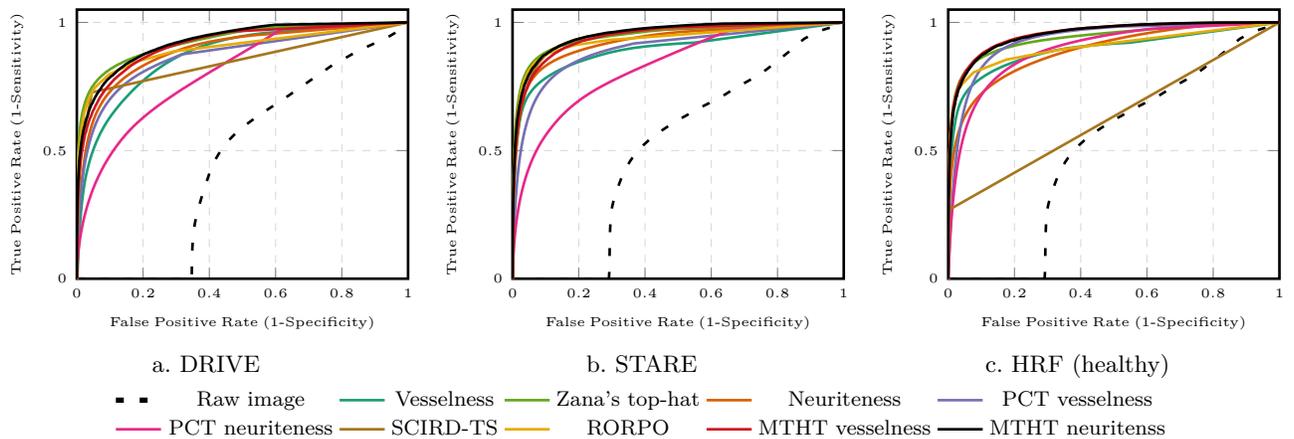
\begin{table*}[h!]
	\centering
	\caption{Mean AUC values for the state-of-the-art approaches and proposed MTHT vesselness and MTHT neuriteness across the DRIVE, STARE and HRF datasets. A section of results are shown in Figure~\ref{fig:rtina} and the mean ROC curves can be seen in Figure~\ref{fig:retinaROC}.}\label{tab:auc2} 
	\resizebox{.75\textwidth}{!}{
		\centering
		\begin{tabular}{lccccccc}
			\toprule
			\multirow{2}{2.5cm}{\centering Enhancement Approach}&\multicolumn{5}{c}{AUC (StDev)}\\ \cmidrule(l){2-4}\cmidrule(l){2-6}
			 &Year/Ref\quad&DRIVE\quad &STARE\quad& HRF (healthy)\quad& HRF (unhealthy)\quad\\
			\midrule
			Raw	image&-											&0.416 (0.064) 			&0.490 (0.076)&0.530 (0.075)& 0.541 (0.073)	\\
			\midrule
			Vesselness&1998~\cite{frangi1998multiscale}			&0.888 (0.243) 			&0.898 (0.215) & 0.913 (0.020)& 0.904 (0.020)\\
			\midrule
		    Zana's top-hat&~2001\cite{zana2001segmentation}		& \hspace{8pt}\textbf{0.933} (0.015) 		&0.956 (0.021) & 0.943 (0.010)& 0.910 (0.016)\\
			\midrule
			Neuriteness&2004~\cite{meijering2004design}	&0.909 (0.022) &0.927 (0.039)& 0.896 (0.024)& 0.879 (0.059)\\
			\midrule
	     	PCT vesselness	&2012~\cite{obara2012contrast}		&0.890 (0.037) 			&0.899 (0.056)			&0.888 (0.011)				&0.837 (0.030)\\
	    	\midrule
		    PCT neuriteness	&2012~\cite{obara2012contrast} 		&0.817 (0.121) 			&0.827 (0.165)			&0.901 (0.029)				&0.777 (0.022)\\
		   \midrule
		   SCIRD-TS&2015~\cite{annunziata2015scale} 			&0.925 (0.468) 			&0.946 (0.021)			&0.956 (0.012)				&0.0.692 (0.035)\\
		   \midrule
			RORPO&2017~\cite{merveille20172d}	&0.867 (0.016) &0.902 (0.020)& 0.869 (0.014)& 0.854 (0.015)\\
			\midrule
			MTHT vesselness &-		&0.923 (0.017) 			&0.955 (0.024)&\textbf{0.959} (0.012)&0.934 (0.015)\\
			\midrule
			MTHT neuritenss	&-		&0.931 (0.016)	&\textbf{0.958} (0.019) &\textbf{0.959} (0.010)&\textbf{0.935} (0.018)\\
			\bottomrule
		\end{tabular}
	}
\end{table*}

\subsection{Application to 2D Retinal Images}
Although a visual inspection can provide some information regarding the effectiveness of the curvilinear structure enhancement approaches, a more rigorous form of quantitative validation is required. 
As in~\cite{jerman2016enhancement}, we chose to use the Receiver Operating Characteristic (ROC) curve and the Area Under the Curve (AUC) metrics to compare the curvilinear structure enhancement approaches. We derive the ROC curve and then calculate the AUC value. Each enhanced image is segmented at different threshold levels and compared with the corresponding ground truth segmentation of curvilinear structures in the image. 
We measure the quality of the approach by using publicly available retinal image datasets:  DRIVE~\cite{niemeijer2004comparative}, STARE~\cite{hoover2000locating} and HRF~\cite{odstrcilik2013retinal}. 
These datasets have been chosen because of their availability and their ground truth data. We have used these ground truth segmentations to quantitatively compare the proposed approach with the other curvilinear structure enhancement approaches.

In particular, we evaluate our approach, alongside the state-of-the-art methods, calculating the Receiver Operating Characteristic (ROC) curve and the mean of Area Under the Curve (AUC) between the enhanced images and the ground truth. 
The results are displayed accordingly in Figure~\ref{fig:rtina}, Figure~\ref{fig:retinaROC} and Table~\ref{tab:auc2}. A higher AUC value indicates a better enhancement of curvilinear structures, with a value of $1$ indicating that the enhanced image is identical to the ground truth image. 
Our experimental results clearly show that our proposed approach works better than the state-of-the-art approaches for the STARE dataset. 
Furthermore, the proposed approach achieved a high score overall on the HRF healthy and unhealthy images, as illustrated in Table~\ref{tab:auc2}. 

The average computation time for the proposed method is $13.7$ seconds for DRIVE image and $16.4$ seconds for STARE image. Please make a note that the proposed method has been implemented and tested in Matlab, however, C++ implementation could be much faster.
\newcommand\hvfigure{2.1cm}
\makeatletter
\define@key{Gin}{mycrops}[]{\setkeys{Gin}{trim={256 120 250 120},clip}}
\makeatother
\newcommand{\bfigure}{0.2}
\begin{figure}[h!]
	\centering	
	\begin{subfigure}[t]{\bfigure\linewidth}
		\includegraphics[mycrops,width=\linewidth,height=\linewidth]{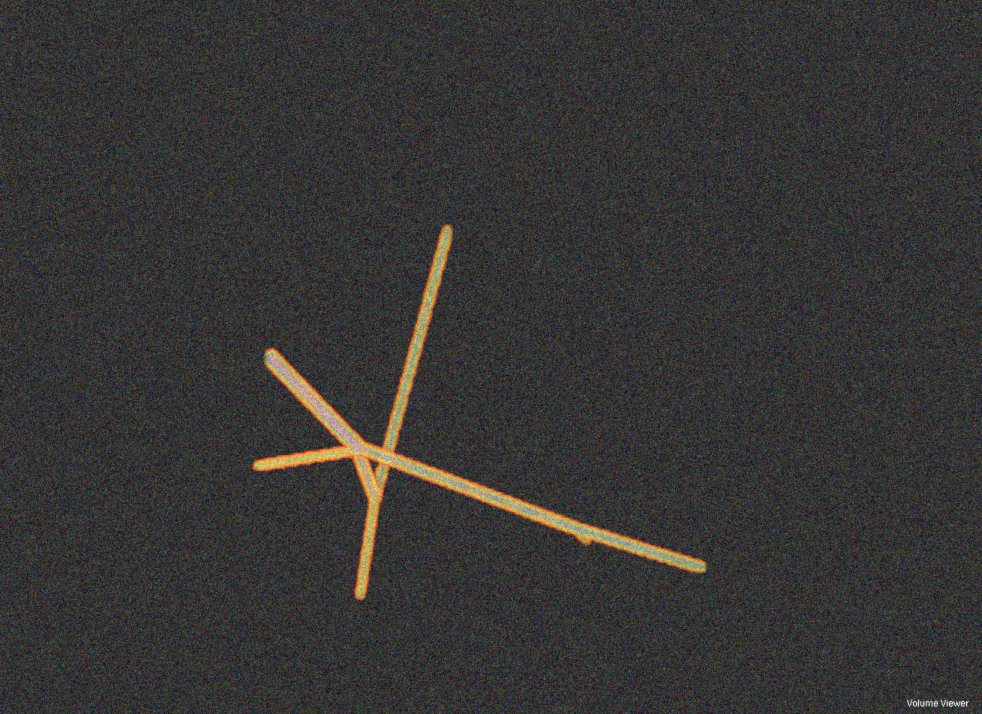}\vspace{4pt}
	\end{subfigure}
	\begin{subfigure}[t]{\bfigure\linewidth}
		\includegraphics[width=\linewidth,height=\linewidth]{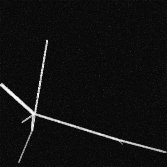}
	\end{subfigure}
	\begin{subfigure}[t]{\bfigure\linewidth}
		\includegraphics[width=\linewidth,height=\linewidth]{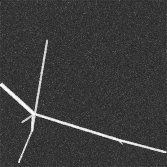}
	\end{subfigure}\\
	\begin{subfigure}[t]{\bfigure\linewidth}
		\includegraphics[mycrops,width=\linewidth,height=\linewidth]{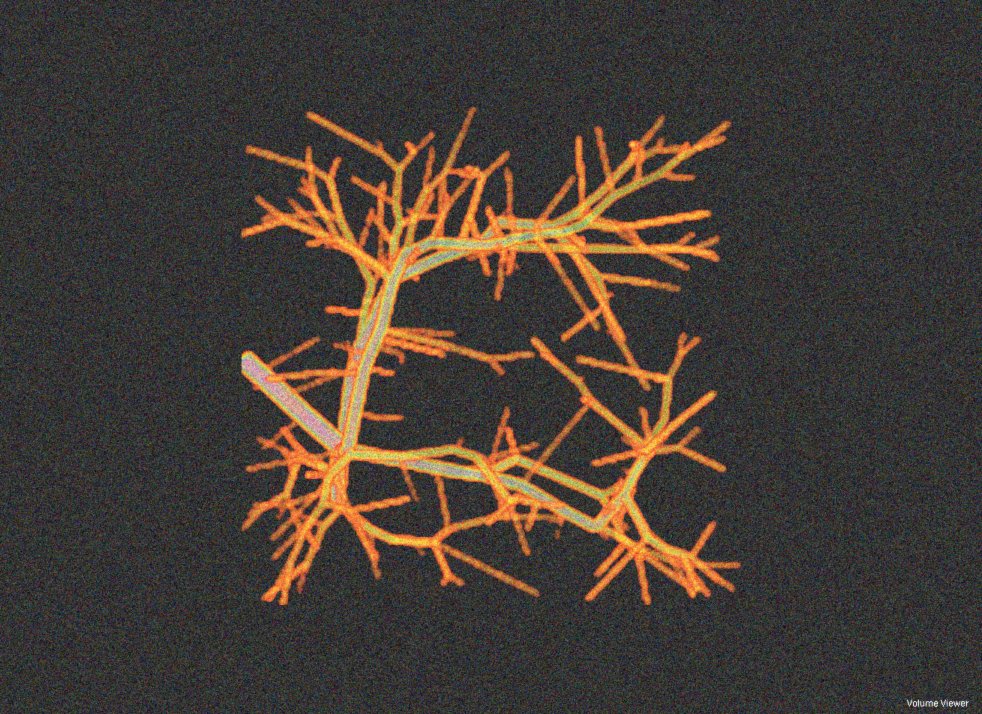}\vspace{4pt}
	\end{subfigure}
	\begin{subfigure}[t]{\bfigure\linewidth}
		\includegraphics[width=\linewidth,height=\linewidth]{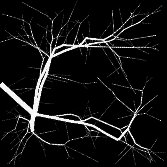}
	\end{subfigure}
	\begin{subfigure}[t]{\bfigure\linewidth}
		\includegraphics[width=\linewidth,height=\linewidth]{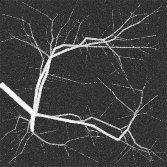}
	\end{subfigure}\\
	\begin{subfigure}[t]{\bfigure\linewidth}
		\includegraphics[mycrops,width=\linewidth,height=\linewidth]{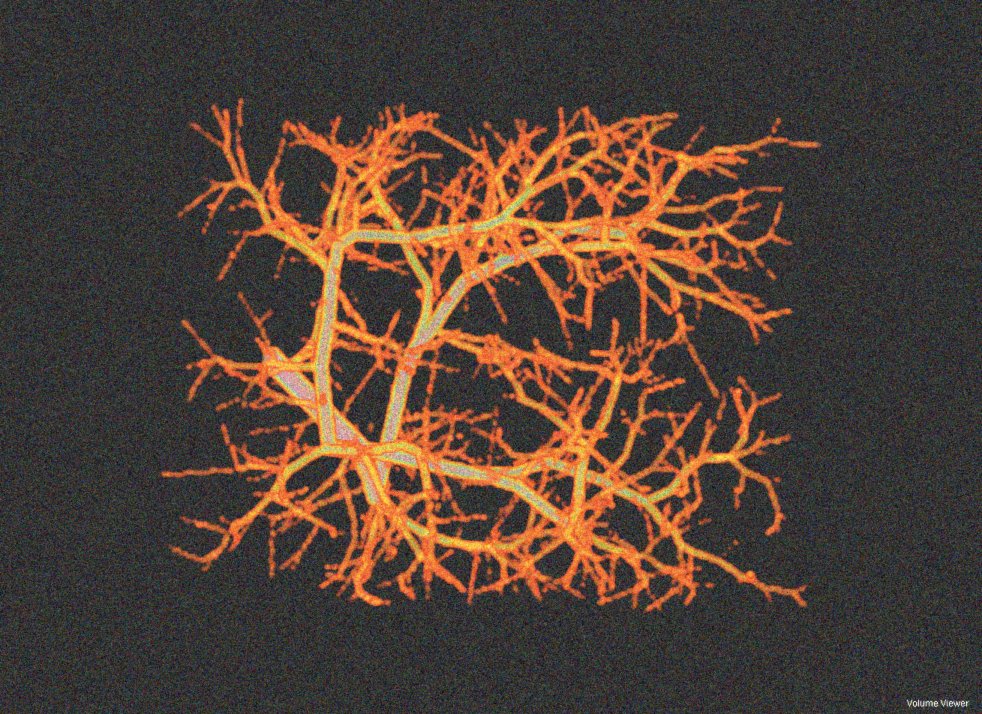}
		\caption{\quad}
	\end{subfigure}
	\begin{subfigure}[t]{\bfigure\linewidth}
		\includegraphics[width=\linewidth,height=\linewidth]{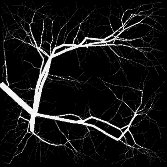}
		\caption{\quad}
	\end{subfigure}
	\begin{subfigure}[t]{\bfigure\linewidth}
		\includegraphics[width=\linewidth,height=\linewidth]{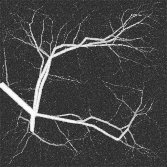}
		\caption{\quad}
	\end{subfigure}
	\caption{A selection of 3D synthetic vascular network images generated with the VascuSynth Software. Each image has a resolution of (167x167x167 voxels) and have different nodes to increase the complexity of structure. (a) original images with different number of nodes (5, 200 and 1000) respectively. (b-c) are the enhance images from the proposed MTHT vesselness and MTHT neuriteness respectively.} \label{fig:vascu:input}
\end{figure}
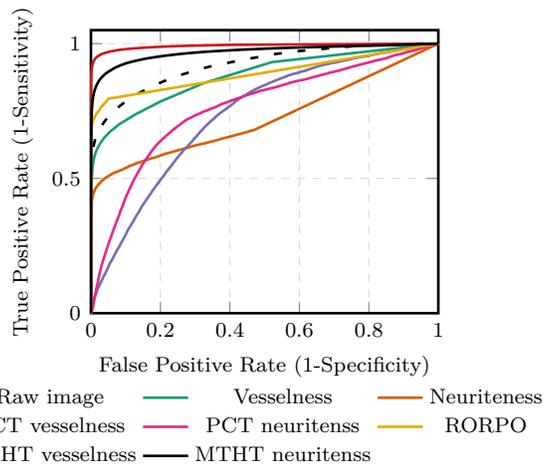
\begin{figure}[h!]
	\begin{subfigure}{\linewidth}
			\centering
		\begin{tikzpicture}
		\begin{axis}[
		width =0.7 \linewidth,
		xlabel={False Positive Rate (1-Specificity)},
		ylabel={True Positive Rate (1-Sensitivity)},
		ylabel near ticks,
		label style={font=\footnotesize},
		tick label style={font=\footnotesize},
		ytickmin=0, ymax=1.05,
		xtickmin=0, xtickmax=1.05,
		enlargelimits=false,
		legend entries = {Raw image, Vesselness, Neuriteness, PCT vesselness, PCT neuritenss, RORPO, MTHT vesselness, MTHT neuritenss},
		legend columns = 3,
		legend style={font=\footnotesize,line width=2pt, draw=none,},
		legend to name=leg3D,
		grid=major, 
		grid style={dashed,gray!30}, 
		]
		\addplot[color=black,loosely dashed] table [x index=0, y index=1, col sep=comma] {images/datFiles/vascuO.dat};
		\addplot[color=brewerDark1,line width=1pt] table [x index=0, y index=1, col sep=comma] {images/datFiles/vascuV.dat};
		\addplot[color=brewerDark2,line width=1pt] table [x index=0, y index=1, col sep=comma] {images/datFiles/vascuN.dat};
		\addplot[color=brewerDark3,line width=1pt] table [x index=0, y index=1, col sep=comma] {images/datFiles/vascuPctV.dat};
		\addplot[color=brewerDark4,line width=1pt] table [x index=0, y index=1, col sep=comma] {images/datFiles/vascuPctN.dat};
        \addplot[color=brewerDark6,line width=1pt] table [x index=0, y index=1, col sep=comma] {images/datFiles/vascuRO.dat};
		\addplot[color=red,line width=1pt] table [x index=0, y index=1, col sep=comma] {images/datFiles/vascuPV.dat};		
		\addplot[color = black,line width=1pt] table [x index=0, y index=1, col sep=comma] {images/datFiles/vascuPN.dat};
		\end{axis}
		\end{tikzpicture}
	\end{subfigure}\quad
    \hspace{5 pt}
	\begin{subfigure}{\linewidth}
		\vskip-3pt
	\centering
	\pgfplotslegendfromname{leg3D}
\end{subfigure}
\caption{Mean ROC curve for all the 9 vascular networks 3D images enhanced using the state-of-the-art approaches alongside the proposed MTHT Vesselness and MTHT Neuiteness (see legend for colours). Correspondingly, the mean AUC values can be found in Table~\ref{tab:auc3}.}\label{fig:roc:vascu}
\end{figure}
\begin{table*}[h!]
\caption{AUC values for the 9 vascular networks 3D images with increasing network's complexity (see Figure~\ref{fig:vascu:input}) enhanced with the state-of-the-art approaches alongside the proposed MTHT vesselness and MTHT neuriteness.}\label{tab:auc3}
	\centering
     \resizebox{.85\textwidth}{!}{
		\begin{tabular}{cccccccc}\toprule
			&\multicolumn{7}{c}{AUC}\\ \cmidrule(l){2-7}\cmidrule(l){2-8}
			{Nodes}	&Vesselness~\cite{frangi1998multiscale}&Neuriteness~\cite{meijering2004design}&PCT vesselness~\cite{sazak2017contrast}&PCT neuritenss~\cite{sazak2017contrast}&RORPO~\cite{merveille2018curvilinear} & MTHT vesselness & MTHT neuritenss\\
			\midrule
			5			&0.999	&0.923& 0.840 &0.897	&0.999 &\textbf{1.000}&0.992\\
			\midrule
			10			&0.996	&0.883& 0.820& 0.873&	0.997 &\textbf{1.000}&0.998\\
			\midrule
			50			&0.976	&0.830& 0.794 &0.851& 0.965 &\textbf{0.999}&0.982\\
			\midrule
			100			&0.951	&0.778&0.778& 0.827& 0.930 &\textbf{0.999}&0.988\\
			\midrule
			200 		&0.930	&0.755& 0.770 &0.799& 0.900 &\textbf{0.998}&0.981\\
			\midrule
			400			&0.910	&0.746& 0.749 &0.788& 0.879 &\textbf{0.996}&0.975\\
			\midrule
			600			&0.902	&0.743& 0.742& 0.777& 0.869 &\textbf{0.993}&0.970\\
			\midrule
			800			&0.885	&0.719& 0.724 &0.756&	0.855 &\textbf{0.987}&0.959\\
			\midrule
			1000		&0.884	&0.722& 0.726 &0.759&	0.852 &\textbf{0.983}&0.956\\
			\midrule
			mean (StDev)	&0.937 (0.045)	&0.788 (0.073)&0.771 (0.040) &0.814 (0.050)& 0.916 (0.058)  &\textbf{0.995} (0.006) &0.978 (0.014)\\ 
			\bottomrule
		\end{tabular}
	}
\end{table*}
\subsection{3D Vascular Network Complexity}
In order to validate our approach in 3D, we used synthetic vascular networks produced by the free software package called VascuSynth ~\cite{hamarneh2010vascusynth}. 
The tree generation is performed by iteratively growing a vascular structure based on an oxygen demand map. Each generated image is associated with it's ground truth. 
In this experiment, we generated 9 volumetric images with an increasing complexity and their corresponding ground truth. 
In addition, in order to make the image more realistic, we added a small amount of Gaussian noise of level $\sigma^2=10$ and applied a Gaussian smoothing kernel with a standard deviation of 1. 
The results, in terms of AUC, are presented in Table~\ref{tab:auc3} and a sample of the results are shown in Figure~\ref{fig:vascu:input}. 
We also demonstrate the mean ROC curve over the 9 enhanced images, as shown in Figure~\ref{fig:roc:vascu}. 
Our proposed approach is compared with vesselness~\cite{frangi1998multiscale}, neuriteness~\cite{meijering2004design}, PCT (vesselness and neuriteness)~\cite{sazak2017contrast} and with the latest 3D enhancement approach~\cite{merveille2018curvilinear}. 
Our proposed approach clearly has the highest mean AUC value (0.995) with a standard deviation equal to (0.006) for the proposed MTHT-vesselness. 
On the other hand, we obtained an AUC value (0.978) with a standard deviation equal to (0.014) for the proposed MTHT-neuriteness compared to the state-of-art approaches.

\subsection{2D and 3D Qualitative Validation}
Additionally, as displayed in Figures~\ref{fig:2d} and~\ref{fig:3d}, we have demonstrated the robustness of the proposed approach when applied to a wide range of 2D and 3D real-world images. 
It is clear that our approach has the best performance compared with the state-of-the-art approaches. 
In particular, our proposed approach can handle complex curvilinear networks as shown in Figure~\ref{fig:3d}(1) and (2).
\newcommand\wfigure{1}
\newcommand\hfigure{1.5cm}
\renewcommand\bfigure{0.09}
\begin{figure*}[h!]
	\centering
	\begin{subfigure}[t]{.008\linewidth}
		\vspace{-25pt}
		1
	\end{subfigure}
	\begin{subfigure}[t]{\bfigure\linewidth}
		\centering
		\begin{tikzpicture}
		\node[anchor=south west,inner sep=0](image)at(0,0)
		{\includegraphics[width=\wfigure\linewidth,height=\hfigure]{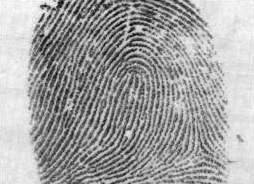}} ;
		\begin{scope}[x={(image.south east)},y={(image.north west)}]
		\draw[-,thick,red] ((0.35,0) rectangle (0.95,.73);
		\end{scope}
		\end{tikzpicture}\vspace{4pt}
	\end{subfigure}
	\begin{subfigure}[t]{\bfigure\linewidth}
		\centering
		\includegraphics[width=\wfigure\linewidth,height=\hfigure]{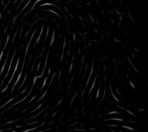}
	\end{subfigure}
		\begin{subfigure}[t]{\bfigure\linewidth}
		\centering
		\includegraphics[width=\wfigure\linewidth,height=\hfigure]{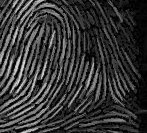}
	\end{subfigure}
	\begin{subfigure}[t]{\bfigure\linewidth}
		\centering
		\includegraphics[width=\wfigure\linewidth,height=\hfigure]{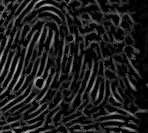}
	\end{subfigure}
	\begin{subfigure}[t]{\bfigure\linewidth}
	\centering
	\includegraphics[width=\wfigure\linewidth,height=\hfigure]{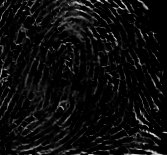}
\end{subfigure}
	\begin{subfigure}[t]{\bfigure\linewidth}
	\centering
	\includegraphics[width=\wfigure\linewidth,height=\hfigure]{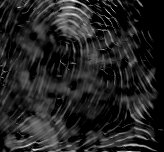}
\end{subfigure}
	\begin{subfigure}[t]{\bfigure\linewidth}
	\centering
	\includegraphics[width=\wfigure\linewidth,height=\hfigure]{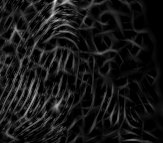}
\end{subfigure}
	\begin{subfigure}[t]{\bfigure\linewidth}
	\centering
	\includegraphics[width=\wfigure\linewidth,height=\hfigure]{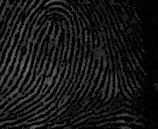}
\end{subfigure}
	\begin{subfigure}[t]{\bfigure\linewidth}
		\centering
		\includegraphics[width=\wfigure\linewidth,height=\hfigure]{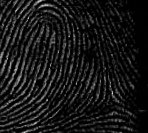}
	\end{subfigure}
	\begin{subfigure}[t]{\bfigure\linewidth}
		\centering
		\includegraphics[width=\wfigure\linewidth,height=\hfigure]{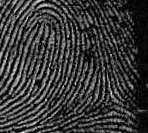}
	\end{subfigure}\\
	\centering
	\begin{subfigure}[t]{.008\linewidth}
		\vspace{-25pt}
		2
	\end{subfigure}
	\begin{subfigure}[t]{\bfigure\linewidth}
		\centering
		\begin{tikzpicture}
		\node[anchor=south west,inner sep=0](image)at(0,0)
		{\includegraphics[width=\wfigure\linewidth,height=\hfigure]{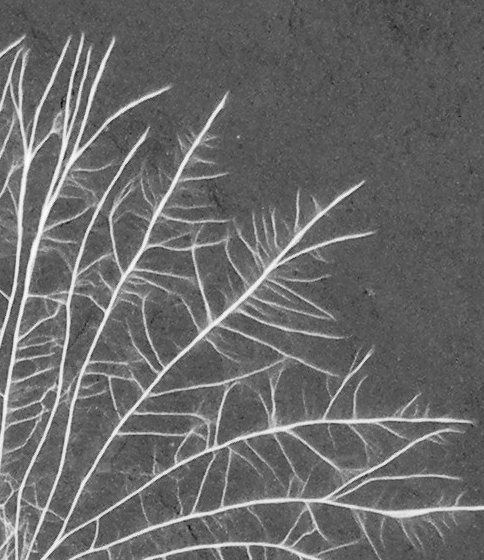}} ;
		\begin{scope}[x={(image.south east)},y={(image.north west)}]
		\draw[-,thick,red] ((0.08,.28) rectangle (0.7,.8);
		\end{scope}
		\end{tikzpicture}\vspace{4pt}
	\end{subfigure}
	\begin{subfigure}[t]{\bfigure\linewidth}
		\centering
		\includegraphics[width=\linewidth,height=\hfigure]{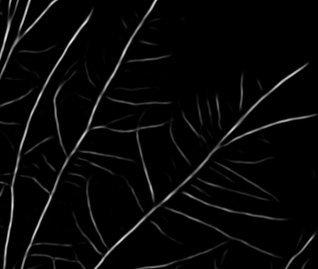}
	\end{subfigure}
			\begin{subfigure}[t]{\bfigure\linewidth}
		\centering
		\includegraphics[width=\wfigure\linewidth,height=\hfigure]{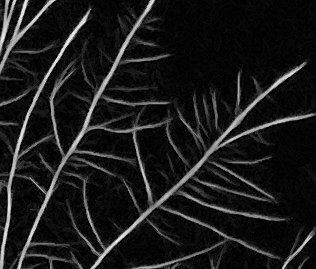}
	\end{subfigure}
	\begin{subfigure}[t]{\bfigure\linewidth}
		\centering
		\includegraphics[width=\linewidth,height=\hfigure]{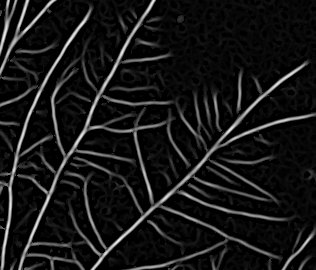}
	\end{subfigure}
	\begin{subfigure}[t]{\bfigure\linewidth}
	\centering
	\includegraphics[width=\linewidth,height=\hfigure]{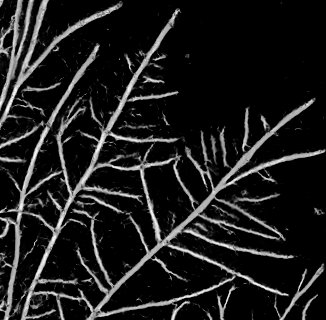}
\end{subfigure}
	\begin{subfigure}[t]{\bfigure\linewidth}
	\centering
	\includegraphics[width=\linewidth,height=\hfigure]{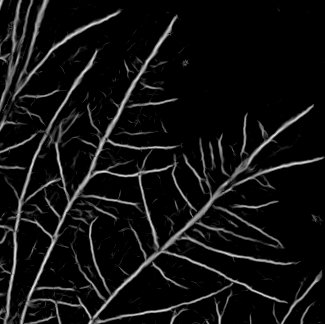}
\end{subfigure}
	\begin{subfigure}[t]{\bfigure\linewidth}
	\centering
	\includegraphics[width=\linewidth,height=\hfigure]{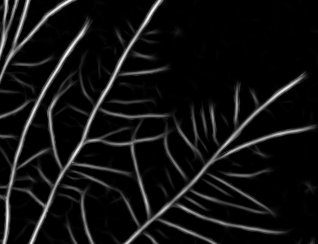}
\end{subfigure}
	\begin{subfigure}[t]{\bfigure\linewidth}
	\centering
	\includegraphics[width=\linewidth,height=\hfigure]{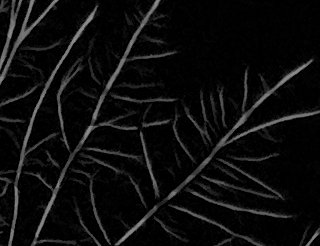}
\end{subfigure}
	\begin{subfigure}[t]{\bfigure\linewidth}
		\centering
		\includegraphics[width=\linewidth,height=\hfigure]{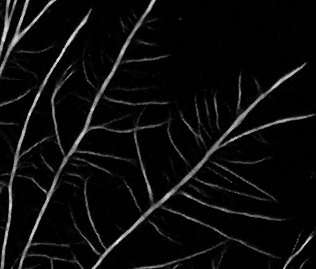}
	\end{subfigure}
	\begin{subfigure}[t]{\bfigure\linewidth}
		\centering
		\includegraphics[width=\wfigure\linewidth,height=\hfigure]{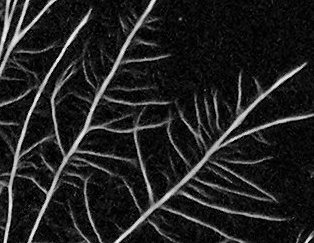}
	\end{subfigure}\\
	\centering
	\begin{subfigure}[t]{.008\linewidth}
	\vspace{-25pt}
		3
	\end{subfigure}
	\begin{subfigure}[t]{\bfigure\linewidth}
		\centering
			\begin{tikzpicture}
		\node[anchor=south west,inner sep=0](image)at(0,0)
		{\includegraphics[width=\wfigure\linewidth,height=\hfigure]{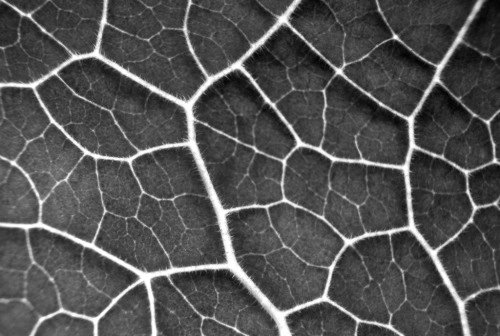}};
		\begin{scope}[x={(image.south east)},y={(image.north west)}]
		\draw[-,thick,red] ((0.05,.35) rectangle (0.7,.95);
		\end{scope}
		\end{tikzpicture}
		\caption{\quad}
	\end{subfigure}
	\begin{subfigure}[t]{\bfigure\linewidth}
		\centering
		\includegraphics[width=\wfigure\linewidth,height=\hfigure]{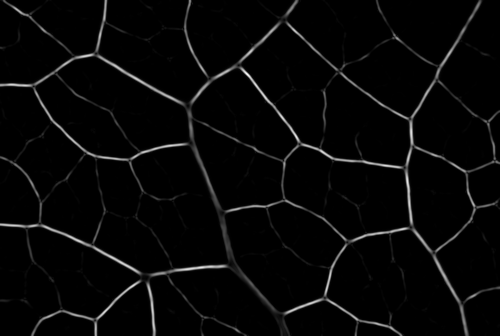}
		\caption{\quad}
	\end{subfigure}
	\begin{subfigure}[t]{\bfigure\linewidth}
	\centering
	\includegraphics[width=\wfigure\linewidth,height=\hfigure]{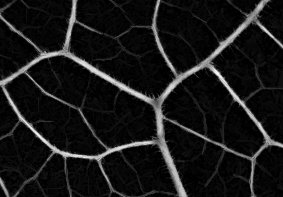}
	\caption{\quad}
\end{subfigure}
	\begin{subfigure}[t]{\bfigure\linewidth}
		\centering
		\includegraphics[width=\wfigure\linewidth,height=\hfigure]{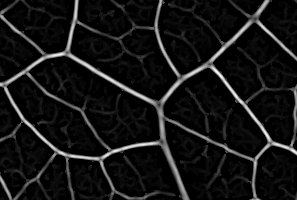}
		\caption{\quad}
	\end{subfigure}
	\begin{subfigure}[t]{\bfigure\linewidth}
	\centering
	\includegraphics[width=\wfigure\linewidth,height=\hfigure]{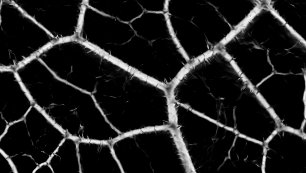}
	\caption{\quad}
\end{subfigure}
\begin{subfigure}[t]{\bfigure\linewidth}
	\centering
	\includegraphics[width=\wfigure\linewidth,height=\hfigure]{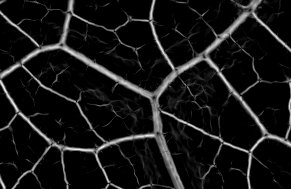}
	\caption{\quad}
\end{subfigure}
\begin{subfigure}[t]{\bfigure\linewidth}
	\centering
	\includegraphics[width=\wfigure\linewidth,height=\hfigure]{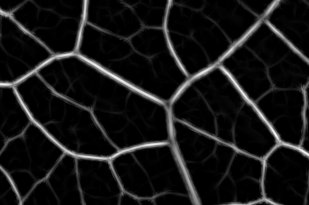}
	\caption{\quad}
\end{subfigure}
	\begin{subfigure}[t]{\bfigure\linewidth}
	\centering
	\includegraphics[width=\wfigure\linewidth,height=\hfigure]{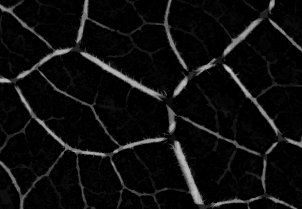}
	\caption{\quad}
\end{subfigure}
	\begin{subfigure}[t]{\bfigure\linewidth}
		\centering
		\includegraphics[width=\wfigure\linewidth,height=\hfigure]{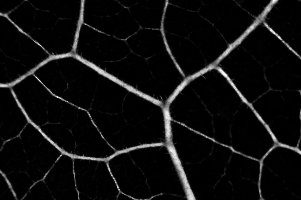}
		\caption{\quad}
	\end{subfigure}
	\begin{subfigure}[t]{\bfigure\linewidth}
		\centering
		\includegraphics[width=\wfigure\linewidth,height=\hfigure]{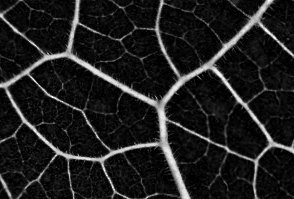}
		\caption{\quad}
	\end{subfigure}
	\caption{Comparison of the curvilinear structure enhancement approaches using 2D real images. (a) original images: (1) finger print \cite{maltoni2009handbook}, (2) macro-scale networks (provided by Prof. M. Fricker, Oxford University, UK), and (3) leaf image \cite{obara2012contrast}. The red box on the original image shows the region of interest. (b) Vesselness~\cite{frangi1998multiscale}, (c) Zana's top-hat~\cite{zana2001segmentation}, (d) Neuriteness~\cite{meijering2004design}, (e) PCT vesselness~\cite{obara2012contrast}, (f) PCT neuriteness~\cite{obara2012contrast}, (g) SCIRD-TS~\cite{annunziata2015scale}, (h) RORPO~\cite{merveille20172d}, (f) MTHT vesselness, and (g) MTHT neuriteness.}
	\label{fig:2d}
\end{figure*}
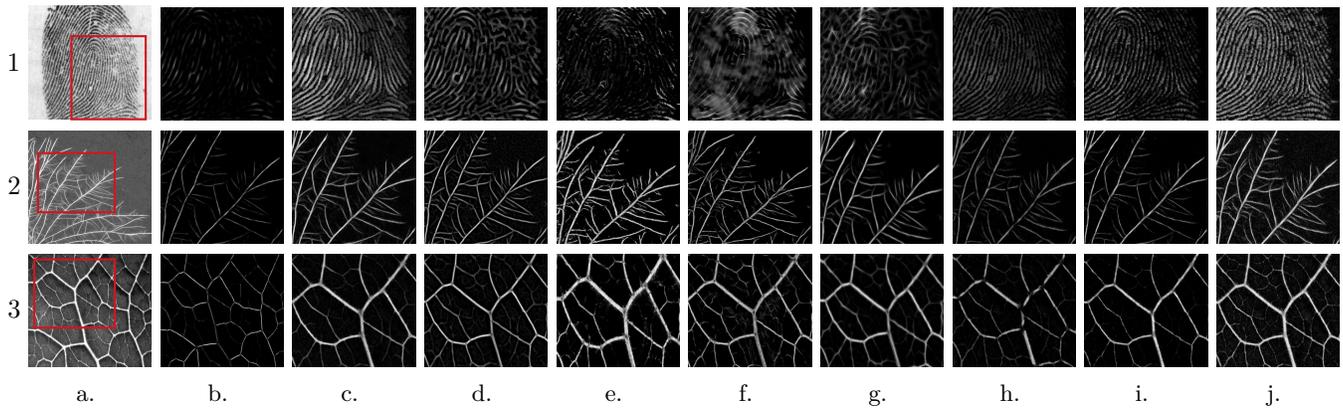

\renewcommand\wfigure{1}
\renewcommand\hfigure{1.25 cm}
\renewcommand\bfigure{0.1}
\begin{figure*}[h!]
	\centering
		\begin{subfigure}[t]{.008\linewidth}
		\vspace{-25pt}
		1
	\end{subfigure}
\begin{subfigure}[t]{\bfigure\linewidth}
	\centering
	\begin{tikzpicture}
	\node[anchor=south west,inner sep=0](image)at(0,0)
	{\includegraphics[width=\wfigure\linewidth,height=\hfigure]{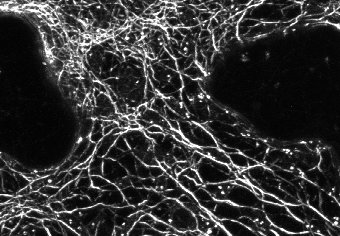}};
	\begin{scope}[x={(image.south east)},y={(image.north west)}]
    \draw[-,thick,red] ((0.29,.12) rectangle (0.95,.73);
	\end{scope}
	\end{tikzpicture}\vspace{4pt}
\end{subfigure}
	\begin{subfigure}[t]{\bfigure\linewidth}
		\includegraphics[width=\wfigure\linewidth,height=\hfigure]{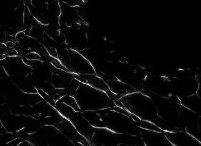}
	\end{subfigure}
	\begin{subfigure}[t]{\bfigure\linewidth}
		\centering
		\includegraphics[width=\wfigure\linewidth,height=\hfigure]{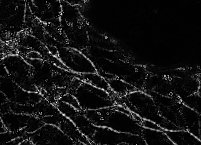}
	\end{subfigure}
	\begin{subfigure}[t]{\bfigure\linewidth}
	\centering
	\includegraphics[width=\wfigure\linewidth,height=\hfigure]{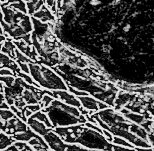}
\end{subfigure}
	\begin{subfigure}[t]{\bfigure\linewidth}
	\centering
	\includegraphics[width=\wfigure\linewidth,height=\hfigure]{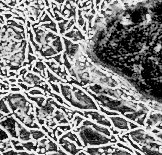}
\end{subfigure}
	\begin{subfigure}[t]{\bfigure\linewidth}
		\centering
		\includegraphics[width=\wfigure\linewidth,height=\hfigure]{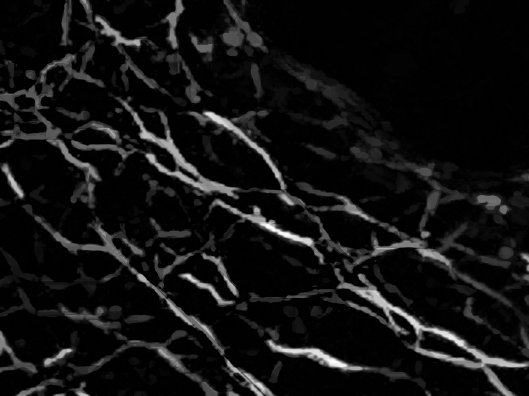}
	\end{subfigure}
	\begin{subfigure}[t]{\bfigure\linewidth}
		\centering
		\includegraphics[width=\wfigure\linewidth,height=\hfigure]{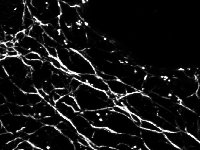}
	\end{subfigure}
	\begin{subfigure}[t]{\bfigure\linewidth}
		\centering
		\includegraphics[width=\wfigure\linewidth,height=\hfigure]{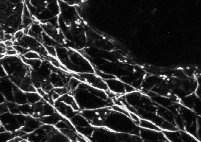}
	\end{subfigure}\\	
	\centering
		\begin{subfigure}[t]{.008\linewidth}
		\vspace{-25pt}
		2
	\end{subfigure}
\begin{subfigure}[t]{\bfigure\linewidth}
	\centering
	\begin{tikzpicture}
	\node[anchor=south west,inner sep=0](image)at(0,0)
	{\includegraphics[width=\wfigure\linewidth,height=\hfigure]{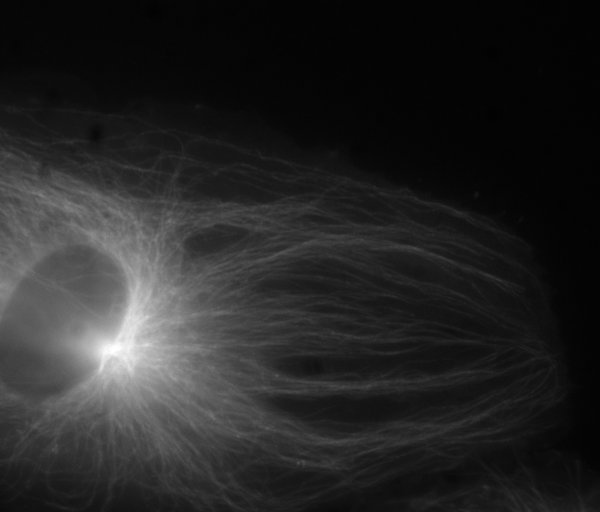}};
	\begin{scope}[x={(image.south east)},y={(image.north west)}]
    \draw[-,thick,red] ((0.17,.12) rectangle (0.8,.73);
	\end{scope}
	\end{tikzpicture}\vspace{4pt}
\end{subfigure}
	\begin{subfigure}[t]{\bfigure\linewidth}
		\centering
		\includegraphics[width=\wfigure\linewidth,height=\hfigure]{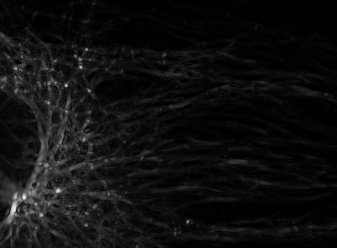}
	\end{subfigure}
	\begin{subfigure}[t]{\bfigure\linewidth}
		\centering
		\includegraphics[width=\wfigure\linewidth,height=\hfigure]{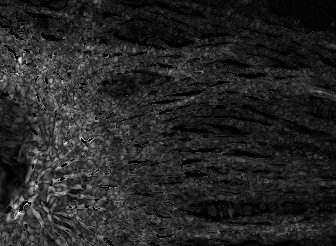}
	\end{subfigure}
	\begin{subfigure}[t]{\bfigure\linewidth}
	\centering
	\includegraphics[width=\wfigure\linewidth,height=\hfigure]{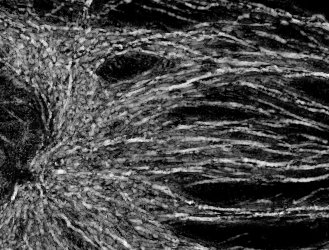}
\end{subfigure}
	\begin{subfigure}[t]{\bfigure\linewidth}
	\centering
	\includegraphics[width=\wfigure\linewidth,height=\hfigure]{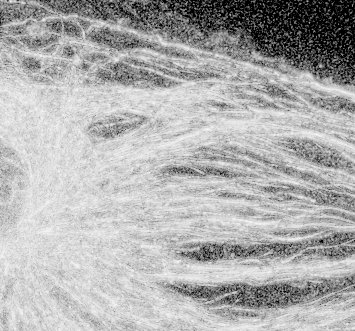}
\end{subfigure}
		\begin{subfigure}[t]{\bfigure\linewidth}
		\centering
		\includegraphics[width=\wfigure\linewidth,height=\hfigure]{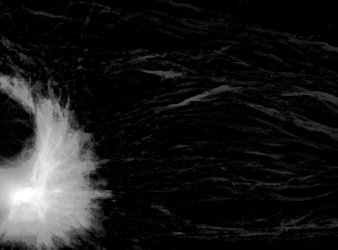}
	\end{subfigure}
	\begin{subfigure}[t]{\bfigure\linewidth}
		\centering
		\includegraphics[width=\wfigure\linewidth,height=\hfigure]{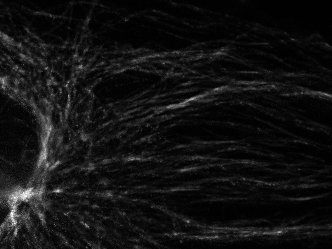}
	\end{subfigure}
	\begin{subfigure}[t]{\bfigure\linewidth}
		\centering
		\includegraphics[width=\wfigure\linewidth,height=\hfigure]{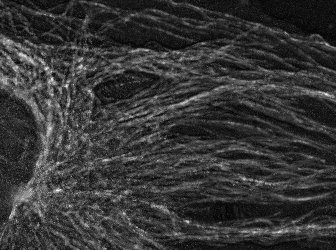}
	\end{subfigure}\\	
	\centering
		\begin{subfigure}[t]{.008\linewidth}
	\vspace{-25pt}
		3
	\end{subfigure}
	\begin{subfigure}[t]{\bfigure\linewidth}
		\begin{tikzpicture}
		\node[anchor=south west,inner sep=0] (image) at (0,0) 
		{\includegraphics[trim={10px 100px 10px 0px}, clip, width=\wfigure\linewidth, height=\hfigure]{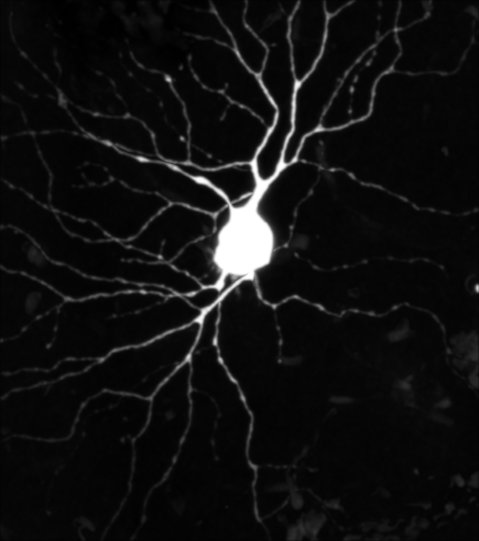}};
		\begin{scope}[x={(image.south east)},y={(image.north west)}]
		\draw[-,thick,red] (0.2,0.38) rectangle (0.8,0.95);
		\end{scope}
		\end{tikzpicture}
		\caption{\quad}
	\end{subfigure}
	\begin{subfigure}[t]{\bfigure\linewidth}
		\centering
		\includegraphics[width=\wfigure\linewidth,height=\hfigure]{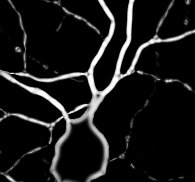}
		\caption{\quad}
	\end{subfigure}
	\begin{subfigure}[t]{\bfigure\linewidth}
		\centering
		\includegraphics[width=\wfigure\linewidth,height=\hfigure]{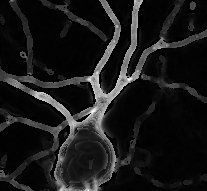}
		\caption{\quad}
	\end{subfigure}
	\begin{subfigure}[t]{\bfigure\linewidth}
	\centering
	\includegraphics[width=\wfigure\linewidth,height=\hfigure]{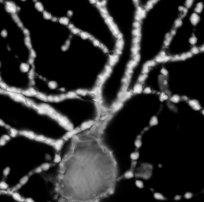}
	\caption{\quad}
\end{subfigure}
\begin{subfigure}[t]{\bfigure\linewidth}
	\centering
	\includegraphics[width=\wfigure\linewidth,height=\hfigure]{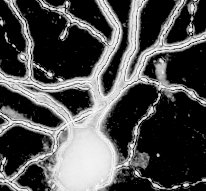}
	\caption{\quad}
\end{subfigure}
		\begin{subfigure}[t]{\bfigure\linewidth}
		\centering
		\includegraphics[width=\wfigure\linewidth,height=\hfigure]{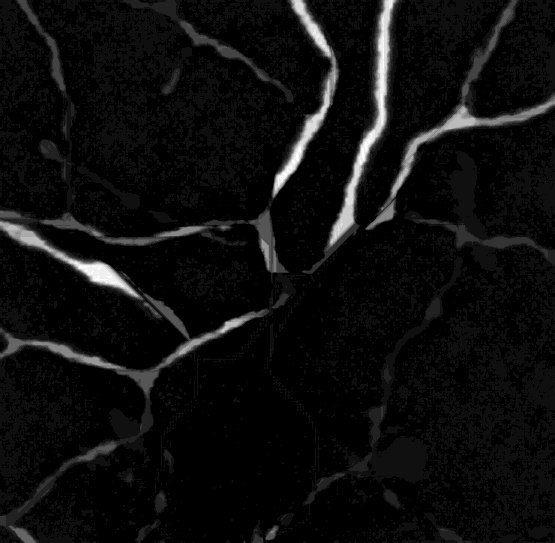}
		\caption{\quad}
	\end{subfigure}
	\begin{subfigure}[t]{\bfigure\linewidth}
		\centering
		\includegraphics[width=\wfigure\linewidth,height=\hfigure]{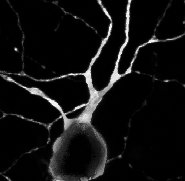}
		\caption{\quad}
	\end{subfigure}
	\begin{subfigure}[t]{\bfigure\linewidth}
		\centering
		\includegraphics[width=\wfigure\linewidth,height=\hfigure]{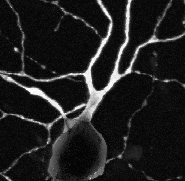}
		\caption{\quad}
	\end{subfigure}\\	
	\caption{Comparison of the curvilinear structure enhancement approaches using 3D real images. (a) original images: (1) keratin network in a skin cell (Dr Tim Hawkins, Durham University, UK), (2) microtubules~\cite{Drelie08-298}, and (3) neuronal (provided by Dr Chris Banna, UC Santa Barbara, USA). The red box on the original image shows the region of interest. (b) Vesselness~\cite{frangi1998multiscale} , (c) Neuriteness~\cite{meijering2004design}, (d) PCT vesselness~\cite{sazak2017contrast}, (e) PCT neuriteness~\cite{sazak2017contrast}, (f) RORPO \cite{merveille2018curvilinear}, (g)  MTHT vesselness, and (h)  MTHT neuriteness.}
	\label{fig:3d}
\end{figure*}
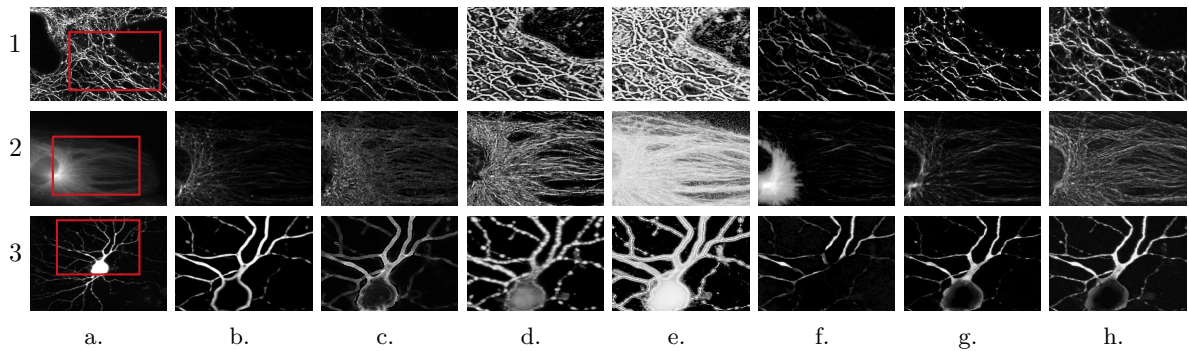
	\section{Implementation}
The software was implemented and written in MATLAB 2017a on Windows 8.1 pro 64-bit PC running an Intel Core i7-4790 CPU (3.60 GHz) with 16GB RAM. The software is made available at: \url{https://github.com/ShuaaAlharbi/MTHT}.

	\section{Conclusion }\label{sec:conclusion}
The enhancement of curvilinear structures is important for many image processing applications. 
In this research, we have proposed a novel approach that combines the advantages of a morphological multiscale top-hat transform and a local tensor to enhance the curvilinear structures in a wide range of 2D and 3D biological and medical images. 

The proposed MTHT approach is evaluated qualitatively and quantitively using different 2D and 3D images. 
The experimental results show that the approach is comparable with the Hessian-based vesselness and neuriteness approaches, as well with the Zana's top-hat, PCT, SCIRD-TS and RORPO approach. 
In general, the MTHT proposed approach showed better enhancement results compared with the state-of-art approaches. 
Although the proposed approach achieves good enhancement results in all tested biomedical images, there is room for improvement. 
In particular, the top-hat transform using different structural elements for an improved enhancement of the image background, as well as better handling of junctions should be explored further.
	\section*{Acknowledgement}
Shuaa Alharbi is supported by the Saudi Arabian Ministry of Higher Education Doctoral Scholarship and Qassim University in Saudi Arabia. \c{C}i\u{g}dem~Sazak is funded by the Turkey Ministry of National Education. Carl J. Nelson is funded by EPSRC UK (EP/N509668/1).

\bibliographystyle{IEEEtran}
\bibliography{bib/Reference}
\end{document}